\documentclass{article} % For LaTeX2e
\usepackage{iclr2023_conference,times}
\iclrfinalcopy
\usepackage{amsmath,amsfonts,bm}

\def\eqref#1{equation~\ref{#1}}
\def\1{\bm{1}}

\DeclareMathAlphabet{\mathsfit}{\encodingdefault}{\sfdefault}{m}{sl}
\SetMathAlphabet{\mathsfit}{bold}{\encodingdefault}{\sfdefault}{bx}{n}

\usepackage{hyperref}
\usepackage{url}
\usepackage{tikz}
\usepackage{comment}
\usepackage{amsmath,amssymb} % define this before the line numbering.
\usepackage{color}
\usepackage{caption}
\usepackage{floatrow}
\usepackage{subcaption}
\usepackage{multirow}
\usepackage{booktabs}
\usepackage{latexsym}
\usepackage[ruled,vlined,linesnumbered]{algorithm2e}
\usepackage{tabulary,multirow,overpic,xcolor}
\usepackage{algorithmic}
\usepackage{wrapfig}

\def\ie{\emph{i.e., }}
\def\eg{\emph{e.g., }}

\newtheorem{theorem}{Theorem}

\title{Maximizing Spatio-Temporal Entropy of Deep 3D CNNs for Efficient Video Recognition}
\author{Junyan Wang$^{1\S}$, Zhenhong Sun$^{12\S}$, Yichen Qian$^2$, Dong Gong$^1$, Xiuyu Sun$^2$\thanks{Corresponding author, $^{\S}$equal contribution, $^{\ddag}$work done in Alibaba.}, Ming Lin$^{3\ddag}$, \\ \textbf{Maurice Pagnucco$^1$, Yang Song$^1$}\\
$^1$University of New South Wales \\
$^2$DAMO Academy, Alibaba Group \\
$^3$Amazon \\
\texttt{\{junyan.wang, dong.gong, yang.song1\}@unsw.edu.au}\\
\texttt{\{zhenhong.szh, yichen.qyc, xiuyu.sxy\}@alibaba-inc.com}\\
\texttt{\{morri\}@cse.unsw.edu.au}, \texttt{\{minglamz\}@amazon.com}}

\begin{document}

\maketitle

\begin{abstract}
% Dong's version:
3D convolution neural networks (CNNs) have been the prevailing option for video recognition. To capture the temporal information, 3D convolutions are computed along the sequences, leading to cubically growing and expensive computations. To reduce the computational cost, previous methods resort to manually designed 3D/2D CNN structures with approximations or automatic search, which sacrifice the modeling ability or make training time-consuming. In this work, we propose to automatically design efficient 3D CNN architectures via a novel training-free neural architecture search approach tailored for 3D CNNs considering the model complexity. To measure the expressiveness of 3D CNNs efficiently, we formulate a 3D CNN as an information system and derive an analytic entropy score, based on the Maximum Entropy Principle. Specifically, we propose a spatio-temporal entropy score (STEntr-Score) with a refinement factor to handle the discrepancy of visual information in spatial and temporal dimensions, 
through dynamically leveraging the correlation between the feature map size and kernel size depth-wisely.
Highly efficient and expressive 3D CNN architectures, \ie entropy-based 3D CNNs (E3D family),  can then be efficiently searched by maximizing the STEntr-Score under a given computational budget, via an evolutionary algorithm without training the network parameters. Extensive experiments on Something-Something V1\&V2 and Kinetics400 demonstrate that the E3D family achieves state-of-the-art performance with higher computational efficiency.
Code is available at \url{https://github.com/alibaba/lightweight-neural-architecture-search}.
\end{abstract}
\vspace{-0.5em}
\section{Introduction}
\vspace{-0.5em}
% \textcolor{red}{backgroud of efficient video recognition}
Video recognition is a fundamental task for video understanding. 
% Over the past decade, there has been increasing research interest along with the emergence of high-quality large-scale video datasets.
% Due to the high complexity of video data, recent research has been focused on proposing high-expressiveness models which, however, also lead to higher computational cost.
To capture the visual information in both temporal and spatial domains from  high-quality large-scale videos, most works have been focusing on proposing highly expressive models which, however, lead to higher computational costs \cite{kondratyuk2021movinets,zhang2022revisiting,liuniformer}.
% Some attempts have been made to design an efficient method by combining a lightweight temporal module with a conventional 2D CNN-based backbone \citep{lin2019tsm,liu2021tam,li2020tea,wang2021tdn,huang2021tada}.
% However, the efficiency of these methods are constrained by the selected 2D CNN-based backbones and the complexity of temporal modules.
Recent research shows that 3D CNNs achieve excellent performance on large-scale benchmarks \citep{hara3dcnns} with unified computations to capture spatio-temporal features jointly. 
% , as 3D CNNs can jointly capture the spatio-temporal features in a unified framework.
% However, the high performance of 3D CNNs comes with a non-cheap price. 
However, the computational cost grows cubically in standard 3D convolution, making it prohibitive for high-resolution long-duration videos.
% \textcolor{red}{challenges of 3D CNN, manually or NAS}
% the reason of why NAS better than non-NAS
Previous works propose to improve the efficiency of 3D CNNs via 2D decomposition or approximation manually \citep{carreira2017i3d,tran2018R21d,feichtenhofer2020x3d}.
Some practices have also been conducted to manually design efficient 3D CNNs relying on heuristics or experiences \citep{hara3dcnns,feichtenhofer2020x3d}. The manually designed 3D or 2D CNN structures cost massive efforts and time in strengthening the modeling ability.
Neural Architecture Search (NAS) approaches \citep{kondratyuk2021movinets,wang2020pv} can automatically generate 3D CNN architectures with higher modeling ability. 
However, searching for a single 3D architecture requires days on multiple GPUs or TPUs, as training and evaluation of the accuracy indicator are required in the process, making the automatic 3D CNN design process time-consuming and/or hardware-dependent.
% Some practices have also been conducted to manually design efficient 3D CNNs relying on heuristics or experiences \cite{hara3dcnns,feichtenhofer2020x3d}. The manually designed 3D or 2D CNN structures cost massive efforts in sacrifice the modeling ability and/or trial-and-error. 
% The manual design not only depends on expert experience, but also costs massive trial-and-error time and computational resources. 
% Thus recent approaches \citep{kondratyuk2021movinets,wang2020pv} are proposed to automatically generate 3D CNN architectures,relying on neural architecture search (NAS) technologies.
% However, these NAS-based approaches rely on the computation-costly training-based accuracy indicator, making the automatic 3D CNN designing process  time-consuming during training time. Typically, hours or even days are required to search for a single neural network fully on multiple GPUs or even TPUs.

To tackle the above issues, we study how to automatically generate (or design) efficient and expressive 3D CNNs with limited computations.
% \textcolor{red}{In application, it is valuable to automatically search an efficient and expressive 3D CNNs within limited computational resource and finite time.}
Recently, training-free technologies have been introduced by some approaches \citep{ntk,lin2021zen,sun2022mae}, in which kernel spectrum analysis or forward inference are adopted to measure the expressiveness of spatial 2D CNNs. 
Inspired by the training-free concept and information theory, we suggest that a deep network can be regarded as an information system, and measuring the expressiveness of the network can be considered equivalent to analyzing how much information it can capture.
Therefore, based on the \textbf{Maximum Entropy Principle} \citep{jaynes1957information}, the probability distribution of the system that best represents the current state of knowledge is the one with the highest entropy.
However, as discussed in~\citep{xie2018s3d}, the information in spatial and temporal domains is different in natural video data.
The spatial dimension is usually limited to some local properties, like connectivity \citep{claramunt2012towards}, while the temporal dimension usually contains more drastic variations with more complex information.
To address the spatio-temporal discrepancy in video data, we conduct a kernel selection experiment and observe that different 3D kernel selections in different stages have different effects on performance, and the focus of 3D CNNs changes from spatial information to spatio-temporal information, as the network depth increases.
We thus consider that the design of 3D CNN architecture should focus on spatial-temporal aggregation depth-wisely.

The above analysis has motivated us to propose a training-free NAS approach to obtain optimal architectures, \ie entropy-based 3D CNNs (\textbf{E3D} family).
Concretely, we first formulate a 3D CNN-based architecture as an information system whose expressiveness can be measured by the value of its differential entropy.
We then derive the upper bound of the differential entropy using an analytic formulation, named \textbf{Spatio-Temporal Entropy Score} (STEntr-Score), conditioned on spatio-temporal aggregation by dynamically measuring the correlation between feature map size and kernel size depth-wisely.
Finally, an evolutionary algorithm is employed to identify the optimal architecture utilizing the STEntr-Score without training network parameters during searching. 
% to evaluate and rank various mutated architectures without training during searching.
In summary, the key contributions of our work are as follows:\\
% $\bullet$ 
% % To the best of our knowledge, 
% We present the \textcolor{blue}{first?novel?} entropy-based training-free neural architecture search approach for automatically designing  efficient 3D CNN architectures. \\
% $\bullet$ The proposed spatio-temporal entropy score, a training-free proxy, with a refinement factor to handle the discrepancy of visual information in spatial and temporal dimensions, dynamically reflects the correlation between the feature map size and kernel size depth-wisely.\\
% $\bullet$ \textcolor{blue}{Each model in E3D family can be generated within less than three hours searching on a 6-core CPU, demonstrating state-of-the-art performance on various datasets under low computational budgets.} at remarkably lower search costs
$\bullet$ We present a novel training-free neural architecture search approach to design efficient 3D CNN architectures. Instead of using forward inference estimation, we calculate the differential entropy of a 3D CNN by an analytic formulation under Maximum Entropy Principle.\\
$\bullet$ We investigate the video data characteristics in spatial and temporal domains  and correlation between feature map with kernel selection, then propose the corresponding spatio-temporal entropy score to estimate the spatio-temporal aggregation dynamically, with a spatio-temporal refinement mechanism to handle the information discrepancy.  \\
$\bullet$ Each model of E3D family can be searched within three hours on a desktop CPU, and the models demonstrate state-of-the-art performance on various video recognition datasets.

\section{Related Work}

% Action recognition has drawn a significant amount of attention from the computer vision community in the past few years.
% Previous methods follow two main directions: temporal structure modeling and 3D convolutional neural networks.
% 2D

\noindent\textbf{Action recognition.}
2D CNNs lack temporal modeling for video sequences, and many approaches \citep{wang2016tsn,lin2019tsm,li2020tea,wang2021tdn,wang2021actionnet,huang2021tada} focused on designing an extended module for temporal information learning.
% TSM \cite{lin2019tsm} firstly introduced temporal modeling to 2D CNN-based frameworks, in which a shift operation for a part of channels was embedded into 2D CNNs.
% Several studies proposed more powerful temporal modules to be embedded into 2D CNN-based backbones. 
% For instance, Li \etal \cite{li2020tea} proposed a motion excitation module and a multiple temporal aggregation module, specifically designed to capture both short- and long-range temporal evolution.
Meanwhile, 3D CNN-based frameworks have a spatio-temporal modeling capability, which improves model performance for video action recognition \citep{tran2015c3d,carreira2017i3d,feichtenhofer2020x3d,kondratyuk2021movinets}.
% For instance, Inflated 3D ConvNet (I3D) \citep{xie2018s3d} is based on 2D ConvNet inflation: filters and pooling kernels of very deep image classification ConvNets are expanded into 3D.
% Recent studies \citep{feichtenhofer2020x3d,kondratyuk2021movinets} demonstrate that 3D CNN-based networks achieve more competitive performance than 2D CNN-based methods but these methods are computationally expensive, both for training and inference.
Some attempts \citep{feichtenhofer2020x3d,fan2020rubiksnet,kondratyuk2021movinets} focused on designing efficient 3D CNN-based architectures.
For example, X3D \citep{feichtenhofer2020x3d} progressively expands a tiny 2D image classification architecture along multiple network axes, in space, time, width and depth.
Our work also focuses on designing efficient 3D CNN-based architectures, but in a deterministic manner with entropy-based information criterion analysis.

\noindent\textbf{Maximum Entropy Principle}.
The Principle of Maximum Entropy is one of the fundamental principles in Physics and Information Theory~\citep{shannon1948,theory1,theory2,theory3}. Accompanied by the widespread applications of deep learning, many theoretical studies \citep{saxe2019information,chan2021redunet,yu2020learning,sun2022mae} try to understand the success of deep learning based on the Maximum Entropy Principle.
% For example, the work of \citep{sun2022mae} maximizes the differential entropy of detection backbones, leading to a better feature extractor for object detection under the same computational budgets. 
% Our work explores the Maximum Entropy Principle in the spatio-temporal task and formulates it with an efficient computation.
Our work focuses on video recognition and explores the aggregation of spatio-temporal information under the Maximum Entropy Principle.

\noindent\textbf{Training-Free NAS}.
To reduce the search time of NAS, recent attempts \citep{naswot,ntk,syn,lin2021zen,sunentropy,sun2022mae,zhou2022training,chen2022nasbenchzero,DBLP:journals/corr/abs-2006-14090} proposed training-free strategies for architecture searching, which construct an alternative score to rank the initialized networks without training.
% SynFlow~\cite{syn} preserves the total flow of synaptic strengths through the network at initialization, subject to a sparsity constraint as the proxy.
% NASWOT~\cite{naswot} computes the architecture score according to the kernel matrix of binary activation patterns between mini-batch samples.
% TE-NAS~\cite{ntk} estimates the expressivity by directly counting the number of active regions on randomly sampled images.
% Zen-NAS~\citep{lin2021zen} proposes a training-free index dubbed Zen-score to rank architectures using the gradient norm of the input image.
For example, the work of \citep{sun2022mae} maximizes the differential entropy of detection backbones, leading to a better feature extractor for object detection under the given computational budgets.
However, these methods construct scores on spatial 2D CNNs, and cannot handle the discrepancy of visual information in spatial and temporal dimensions of 3D CNNs.
In order to address the above issues, our work aims to optimize the network architecture by considering spatio-temporal dimensions aggregation.

% \newpage

\section{The Proposed Approach}
In this section, we first present a detailed technical description of the derivation process of an analytical solution and propose the STEntr-Score with a refinement factor to handle the discrepancy of visual information in spatial and temporal dimensions.
Then we give an overview of the search strategy for the E3D family, via an evolutionary algorithm without training the network parameters.
% In this section, we first present a detailed technical description of the derivation process of the MaxE score.
% Then, we give an overview of our MaxE3D architecutre design strategy and insights.
% This section describes the derivation process of the MaxE score and design strategy of MaxE3D for a spatio-temporal information system.
% We also give our detailed answer for kernel size selection and channel-wise expansion in detail which introduced in Section 1.

\subsection{Preliminary}
In \textit{Information Theory}, differential entropy is employed to represent the information capacity of an information system by measuring the output of the system~\citep{jaynes1957information,theory1,theory2,theory3,entropy}.
Generally, the output of a system is a high-dimensional continuous variable with a complex probability distribution, making it difficult to compute the precise value of its entropy directly.
Based on the \textbf{Maximum Entropy Principle} \citep{jaynes1957information}, a common alternative approach is to estimate the upper bound of the entropy \citep{Elements2012}, as:
\begin{theorem}
For any continuous distribution $P(x)$ of mean $\mu$ and variance $\sigma^2$, its differential entropy is maximized when $P(x)$ is a Gaussian distribution $\mathcal{N}(\mu,\sigma^2)$.
\label{trm:gaussian}
\end{theorem}
Thus, the differential entropy of any distribution is upper bound by the Gaussian distribution with the same mean and variance.
Suppose $x$ is sampled from Gaussian distribution $\mathcal{N} (\mu, \sigma^2)$, the differential entropy~\citep{entropy} of $x$ is then:
% equation
\begin{equation}
    % \begin{aligned}
  \mathcal{H}(x) = \int_{-\infty}^{+\infty}-log(P(x))P(x)dx ~~\propto  log(\sigma^2),
      % &= (\frac{1}{2}log(2\pi) + log(\sigma))\int_{-\infty}^{+\infty}p(x)dx + \int_{-\infty}^{+\infty}\frac{(x-\mu)^2}{2\sigma^2}p(x)dx\\
    %   = \frac{1}{2}log(2\pi{e}\sigma^2)  ,\\
  \label{eq:gaussian}
    % \end{aligned}
\end{equation}
where $P(x)$ represents the probability density function of $x$.
Note that the entropy of the Gaussian distribution depends only on the variance $\sigma^2$, and a simple proof is included in the \textbf{Appendix \ref{ssec:proof entropy}}.

According to successful deep learning applications~\citep{saxe2019information,chan2021redunet,yu2020learning,sun2022mae,sunentropy} of Maximum Entropy Principle, a deep neural network can be regarded as an information system, and the differential entropy of the last output feature map represents the expressiveness of the system.
Recent method \citep{sun2022mae} estimates the entropy of 2D CNNs by simply computing the feature map variance via sampling input data and initializing network parameters from a random standard Gaussian distribution.
% \textcolor{red}{must explain the difference in introduction or related work.}
However, when migrating to 3D CNNs, 
how to efficiently reduce the random sampling noise due to the random initialization, and how to estimate the entropy after aggregating spatial and temporal dimensions in 3D CNNs design, still remain open questions. 
We then propose our method to address these problems.

\subsection{Statistical Analysis of Entropy in Deep 3D CNNs}
% \subsection{\textcolor{red}{Maximum Entropy Principle in Deep 3D CNNs}}
% The input video data ${\boldsymbol {x}}^0$ is generated by a standard Gaussian distribution and the forward inference of the network is performed to compute the variance of ${\boldsymbol {x}}^L$.
% To evaluate the information capacity of the spatio-temporal system, we need to compute the variance of ${\boldsymbol {x}}^L$.
% \textcolor{red}{Previous works \cite{lin2021zen,sun2022mae} random sampling input data from the Gaussian distribution, there might exist variance between different samples on the same network, which might lead to an inconsistent differential entropy. 
% Conventional methods for removing random sampling noise are based on increasing the number of samples, such as a big batch size and large numbers of iterations. However, these operations will cost higher computational resources.  
% To this end, we aim to optimize this forward inference with an alternative stable value.}

\noindent\textbf{Simple Network Space}.
% \textcolor{blue}{To conduct analysis of network architectures, vanilla 3D CNN networks without considering auxiliary modules (e.g., BN, Reslink, SE and so on.) are applied in our system design, which follows the customary rules in~\citep{lin2021zen,sun2022mae}.
% % and we aim to build an effective entropy evaluation to estimate the network performance.
% Since these auxiliary modules are unfavorable for the analysis of the system, they are ignored during entropy computing and are plugged into the backbone without special modification during training. 
% Additionally, for holistic analysis, the bias of the convolutional layer is set to zero and activation function is omitted for simplification.
% A detailed discussion about customary rules is included in \textbf{Appendix \ref{sec:simple network}}.}
Following the idea that \textit{simpler is better} \citep{lin2021zen,sun2022mae}, we apply vanilla 3D CNNs without considering auxiliary modules (\eg BN \citep{ioffe2015batch}, Reslink \citep{he2016deep}, SE block \citep{hu2018squeeze} and so on) to conduct analysis of network architectures.
Formally, given a convolutional network with $L$ layers of weights ${\boldsymbol{W}}^1$,  $...$,  ${\boldsymbol{W}}^L$, the forward inference with a  simple network space is given by:
\begin{equation}
{\boldsymbol {x}}^{l} = {{\boldsymbol{W}}^l} * {\boldsymbol {x}}^{l-1} \quad \text{ for } l=1, \dots, L \, ,
\label{eq:forward}
\end{equation}
where $x^l$ denotes the $l^{th}$ layer feature map. For holistic analysis, the bias of the convolutional layer is set to zero and the activation function is omitted in the network for simplification.
Auxiliary modules are ignored during entropy calculation and plugged into the backbone without special modification during training.
A detailed discussion about these rules is included in \textbf{Appendix \ref{sec:simple network}}.
% \textcolor{blue}{lack of connection.}
% \textcolor{blue}{To conduct the forward inference of calculating entropy, random distribution for sampling input data and initializing network parameters are needed~\citep{sun2022mae}.
% Due to the random initialization, the forward entropy is insistent during different sampling, leading to sampling noise. For attenuating sampling noise, the conventional methods are computing average value by multiplying sampling number, and increasing the value of batch size or resolution. 
% Unfortunately, these operations will cost higher time-consuming and computational resources, which is unaffordable in 3D CNNs. 
% Therefore, we will optimize the forward inference with an alternative stable value in a mathematical way.}
% 后面两句应该整合成一个长句会更简洁
% \textcolor{blue}{To conduct forward inference of calculating entropy, input data and network parameters are random sampled from Gaussian Distribution \citep{sun2022mae}.}
% To conduct forward inference of calculating entropy, initialization of input data and network parameters are required by random sampling from Gaussian Distribution \citep{sun2022mae}.

Since the input data and network parameters are randomly sampled from Gaussian distributions, the forward entropy calculation will be inconsistent, which might lead to random sampling noise.
% Since the random initialization is adopted, \textcolor{blue}{the forward entropy calculation will be inconsistent}, which might lead to random sampling noise.
To obtain a valid entropy value, computing an average value from multiple sampling iterations and increasing the value of batch size or resolution can be adopted to reduce the noise. 
% Conventional methods for \textcolor{blue}{, eliminating?attenuating} random sampling noise include computing average value from multiple sampling iterations and increasing the value of batch size or resolution.
These operations are however time-consuming and cost higher computational resources.
% To this end, we will optimize the forward inference with an alternative stable value.
To this end, we propose to explore the statistical characteristics of the forward inference, to provide an efficient solution. 
% \textcolor{blue}{?lack of connection, why we can.}
% The stable value won't

\noindent\textbf{Maximum Entropy of 3D CNNs}.
% \noindent\textbf{Expectation and Variance}.
We first consider the \textit{product law of expectation} \citep{mood1950introduction} and the \textit{Bienaymé's identity} in probability theory \citep{loeve2017probability}, as follows:
\begin{theorem}
Given two independent random variables $v_1$, $v_2$, the expectation of their product $v_1v_2$ is: $\mathbb{E}(v_1v_2)=\mathbb{E}(v_1)\mathbb{E}(v_2)$.
\label{trm:product}
\end{theorem}
\begin{theorem}
Given $n$ random variables $\{v_1, v_2, ..., v_i, v_{i+1}, ..., v_n\}$ which are pairwise independent integrable, the sums of their expectations and variances are: $\mathbb{E}(\sum_{i=1}^n v_i) = \sum_{i=1}^n {\mathbb{E}}(v_i)$, and $\mathbb{D}^2(\sum_{i=1}^n v_i) = \sum_{i=1}^n {\mathbb{D}^2}(v_i)$.
\label{trm:sum}
\end{theorem}
% \begin{equation}
%     E(XY)=E(X)E(Y), 
%     \label{eq:product_expect}
% \end{equation} 
% page 160 of Introduction to the Theory of Statistics by Mood-Graybill-Boes, 3rd edition
We can thus compute the expectation and variance of $l^{th}$ layer feature map element ${\boldsymbol {x}}^l_{i}$ as:
% \begin{equation}
%     \mathbb{D}^2(v_1v_2) = \mathbb{D}^2(v_1)\mathbb{D}^2(v_2)+\mathbb{D}^2(v_2){[\mathbb{E}(v_1)]^2}+\mathbb{D}^2(v_1){[\mathbb{E}(v_2)]^2}~,
%     \label{eq:product_variance}
% \end{equation} %
\begin{equation}
\small
    % \mathbb{E}({\boldsymbol {x}}^l_{i}) = \mathbb{E}(\sum_{t=1}^{K^{l}_t} \sum_{h=1}^{K^{l}_h} \sum_{w=1}^{K^{l}_w} \sum_{c=1}^{C^{l-1}} {\boldsymbol {x}}^{l-1}_{cthw} {\boldsymbol{W}}^l_{cthw}) = \sum_{t=1}^{K^{l}_t} \sum_{h=1}^{K^{l}_h} \sum_{w=1}^{K^{l}_w} \sum_{c=1}^{C^{l-1}} \Big[\mathbb{E}({\boldsymbol {x}}^{l-1}_{cthw}) \mathbb{E}({\boldsymbol{W}}^l_{cthw})\Big],
    \mathbb{E}({\boldsymbol {x}}^l_{i}) = \sum_{t=1}^{K^{l}_t} \sum_{h=1}^{K^{l}_h} \sum_{w=1}^{K^{l}_w} \sum_{c=1}^{C^{l-1}} \Big[\mathbb{E}({\boldsymbol {x}}^{l-1}_{cthw}) \mathbb{E}({\boldsymbol{W}}^l_{cthw})\Big],
    % &={K^{l}_t}{K^{l}_h}{K^{l}_w}C^{l-1}\left[E({\boldsymbol{x}}^{l-1}_{cthw}{\boldsymbol{W}}^l_{cthw})\right]\\
    % &={K^{l}_t}{K^{l}_h}{K^{l}_w}C^{l-1}\left[E({\boldsymbol{x}}^{l-1}_{cthw})\times E({\boldsymbol{W}}^l_{cthw})\right],
    % &={K^{l}_t}{K^{l}_h}{K^{l}_w}C^{l-1}Var({\boldsymbol {x}}^{l-1}_{cthw}],\  with\ Var[{\bf {W}}^l_{cthw}]=1, 
    \label{eq:expectation}
\end{equation} % 
% \begin{equation}
% \small
%     \begin{split}
%     \mathbb{D}^2({\boldsymbol {x}}^l_{i}) &= \mathbb{D}^2(\sum_{t=1}^{K^{l}_t} \sum_{h=1}^{K^{l}_h} \sum_{w=1}^{K^{l}_w} \sum_{c=1}^{C^{l-1}} {\boldsymbol {x}}^{l-1}_{cthw} {\boldsymbol{W}}^l_{cthw})
%     %  = \sum_{t=1}^{K^{l}_t} \sum_{h=1}^{K^{l}_h} \sum_{w=1}^{K^{l}_w} \sum_{c=1}^{C^{l-1}} {\mathbb{D}^2(\boldsymbol {x}}^{l-1}_{cthw} {\boldsymbol{W}}^l_{cthw})\\
%      =\sum_{t=1}^{K^{l}_t} \sum_{h=1}^{K^{l}_h} \sum_{w=1}^{K^{l}_w} \sum_{c=1}^{C^{l-1}}\Big\{\mathbb{D}^2({\boldsymbol{x}}^{l-1}_{cthw}) \mathbb{D}^2({\boldsymbol{W}}^l_{cthw})\\
%     &\quad+\mathbb{D}^2({\boldsymbol{x}}^{l-1}_{cthw}){\Big[\mathbb{E}({\boldsymbol{W}}^l_{cthw})\Big]^2}+\mathbb{D}^2({\boldsymbol{W}}^l_{cthw}){\Big[\mathbb{E}({\boldsymbol{x}}^{l-1}_{cthw})\Big]^2}\Big\},\\
%     % &={K^{l}_t}{K^{l}_h}{K^{l}_w}C^{l-1}\left[\sigma^2({\boldsymbol{x}}^{l-1}_{cthw}{\boldsymbol{W}}^l_{cthw})\right]\\
%     % &=\sum_{t=1}^{K^{l}_t} \sum_{h=1}^{K^{l}_h} \sum_{w=1}^{K^{l}_w} \sum_{c=1}^{C^{l-1}}\sigma^2({\boldsymbol {x}}^{l-1}_{cthw}).\\
%     % &={K^{l}_t}{K^{l}_h}{K^{l}_w}C^{l-1}\sigma^2({\boldsymbol {x}}^{l-1}_{cthw}).
%     % &={K^{l}_t}{K^{l}_h}{K^{l}_w}C^{l-1}Var({\boldsymbol {x}}^{l-1}_{cthw}],\  with\ Var[{\bf {W}}^l_{cthw}]=1, 
%     \end{split}
%     \label{eq:variance}
% \end{equation}
\begin{equation}
\small
    \begin{split}
    \mathbb{D}^2({\boldsymbol {x}}^l_{i})=\sum_{t=1}^{K^{l}_t} \sum_{h=1}^{K^{l}_h} \sum_{w=1}^{K^{l}_w} \sum_{c=1}^{C^{l-1}}&\Big\{\mathbb{D}^2({\boldsymbol{x}}^{l-1}_{cthw}) \mathbb{D}^2({\boldsymbol{W}}^l_{cthw})+\mathbb{D}^2({\boldsymbol{x}}^{l-1}_{cthw}){\Big[\mathbb{E}({\boldsymbol{W}}^l_{cthw})\Big]^2}\\
    &+\mathbb{D}^2({\boldsymbol{W}}^l_{cthw}){\Big[\mathbb{E}({\boldsymbol{x}}^{l-1}_{cthw})\Big]^2}\Big\},\\
    % &={K^{l}_t}{K^{l}_h}{K^{l}_w}C^{l-1}\left[\sigma^2({\boldsymbol{x}}^{l-1}_{cthw}{\boldsymbol{W}}^l_{cthw})\right]\\
    % &=\sum_{t=1}^{K^{l}_t} \sum_{h=1}^{K^{l}_h} \sum_{w=1}^{K^{l}_w} \sum_{c=1}^{C^{l-1}}\sigma^2({\boldsymbol {x}}^{l-1}_{cthw}).\\
    % &={K^{l}_t}{K^{l}_h}{K^{l}_w}C^{l-1}\sigma^2({\boldsymbol {x}}^{l-1}_{cthw}).
    % &={K^{l}_t}{K^{l}_h}{K^{l}_w}C^{l-1}Var({\boldsymbol {x}}^{l-1}_{cthw}],\  with\ Var[{\bf {W}}^l_{cthw}]=1, 
    \end{split}
    \label{eq:variance}
\end{equation}
where $\{{K^{l}_t},{K^{l}_h},{K^{l}_w}\}$ represents the kernel size of the $l^{th}$ layer in the 3D CNN, and $C^{l-1}$ denotes its input channels size. Note that $C^{l-1}$ is equal to 1 when the layer is a depth-wise convolution.
Besides, $t,h,w$ denote the temporal, height, and width positions, respectively.
A simple proof is included in \textbf{Appendix \ref{ssec:proof variance}}.
The input $x^0$ is initialized from a standard Gaussian distribution, which means that its expectation $\mathbb{E}(x^0)=0$ and variance $\mathbb{D}^2(x^0)= 1$.
From the perspective of statistics, we can regard $\mathbb{D}^2(x^0_{cthw}) = 1$ when sampling sufficient times.
% \begin{equation}
%     \mathbb{E}({\boldsymbol {x}}^{1}_{i}) =0, \quad \mathbb{D}^2({\boldsymbol {x}}^{1}_{i}) =  \sum_{t=1}^{K^{l}_t}\sum_{h=1}^{K^{l}_h} \sum_{w=1}^{K^{l}_w} \sum_{c=1}^{C^{l-1}} \Big\{\mathbb{D}^2(\boldsymbol{W}^1_{chw}) + \Big[\mathbb{E}({\boldsymbol{W}^1_{chw}})\Big]^2\Big\}.
%     \label{eq:next}
% \end{equation}
% Also, suppose that all parameters are initialized from a Gaussian distribution $\mathcal{N}(0, \sigma^2_w)$, and thus the variance ${\boldsymbol {x}}^{1}_{i}$ is derived as:
% \begin{equation}
%     \mathbb{D}^2({\boldsymbol {x}}^1_{i}) = \sum_{t=1}^{K^{l}_t}\sum_{h=1}^{K^{l}_h} \sum_{w=1}^{K^{l}_w} \sum_{c=1}^{C^{l-1}} \mathbb{D}^2({\boldsymbol{W}^1_{chw}}) .
%     \label{eq:x_d}
% \end{equation}
Also, suppose that all parameters are initialized from a zero-mean Gaussian distribution, and thus the variance of the last layer $\mathbb{D}^2(x^L)$ can be computed by propagating the variances from previous layers as:
\begin{equation}
    % \begin{split}
    % Var({\boldsymbol {x}}^L_{cthw}) &= \prod^{L}_{l=1}{{K^{l}_t}{K^{l}_h}{K^{l}_w}C^{l-1}}Var({\boldsymbol {x}}^{0}_{cthw})\\
    % &= \prod^{L}_{l=1}{{K^{l}_t}{K^{l}_h}{K^{l}_w}C^{l-1}}, ~~ \text{where} ~Var({\boldsymbol {x}}^{0}_{cthw})=1.
    % \end{split}
    \mathbb{D}^2({\boldsymbol {x}}^L_{i}) = \prod^{L}_{l=1}{{K^{l}_t}{K^{l}_h}{K^{l}_w}C^{l-1}}\mathbb{D}^2({\boldsymbol{W}^l_{cthw}}).
    \label{eq:x_d}
\end{equation}
Finally, by combining Eq. (\ref{eq:x_d}) and Eq. (\ref{eq:gaussian}), we derive that the upper bound entropy is numerically proportional to:
\begin{equation}
    \mathcal{H}(F)  \propto  \sum^{L}_{l=1}log({{K^{l}_t}{K^{l}_h}{K^{l}_w}C^{l-1}}\mathbb{D}^2({\boldsymbol{W}^l_{cthw}}))~,\\
    \label{eq:maxe_score}
\end{equation}
where detailed proof is included in \textbf{Appendix \ref{ssec:proof 3d cnn}}. 
By assuming that the parameters of each layer are initialized with a standard Gaussian distribution with $\mathbb{D}({\boldsymbol{W}^l_{cthw}})=1$, the entropy score defined in Eq. (\ref{eq:maxe_score}) can be written as  $\sum^{L}_{l=1}log({{K^{l}_t}{K^{l}_h}{K^{l}_w}C^{l-1}})$. It measures the influence of kernel size and channel dimension on the entropy score in a homogeneous way, named \textbf{HomoEntr-Score}. 
% When parameters of each layer are also initialized from standard Gaussian Distribution, 
% the calculation of entropy in Eq. \ref{eq:maxe_score} only consists of kernel size and channel dimension, nameed \textbf{HomoEntr-Score}.
This analytic formulation does not require random sampling, thus no random sampling noise exists.

\subsection{Spatio-temporal Entropy Score}
The HomoEntr-Score is derived from the analysis with an independent and identical assumption on the input elements (and the corresponding intermediate features). Although it can generally represent the expressiveness characteristics of a neural network, there is a gap between HomoEntr-Score and reality on 3D CNNs. 
When directly applying it on 3D CNNs for handling video sequences, we realize the HomoEntr-Score with the independent and identical assumption cannot capture the discrepancy of the visual information in the spatial and temporal domain, as the information between spatial and temporal dimensions in video data is different in video recognition. 
% However in natural video data, the information between spatial and temporal dimensions is different in action recognition task \citep{xie2018s3d}. 
% The spatial dimension is usually limited to some local properties, like connectivity \citep{claramunt2012towards}, while the temporal dimension usually contains more drastic variations with more complex information. 
The gap 
% between the assumption in the design of HomoEntr-Score and the video data characteristics 
leads to some issues with HomoEntr-Score for modeling video data with 3D CNNs. The observations will be discussed and analyzed in the following. 
Note that HomoEntr-Score (and similar approaches \citep{sun2022mae}) can work well for modeling the expressiveness of 2D CNNs since there is no (obvious) discrepancy on the information of the two directions in 2D images statistically. 
Based on the analyses, we propose a Spatio-Temporal Entropy Score (STEntr-Score) for 3D CNNs on video data, where a Spatio-temporal refinement factor is introduced to handle the information discrepancy.

\begin{table}[h]
\caption{Results of different kernel positions on the Sth-Sth V1 validation dataset. All model structures are based on X3D-S \citep{feichtenhofer2020x3d}. ``S-N" models mean only stage N selects 1$\times$5$\times$5 kernel, and others select 3$\times$3$\times$3. ``T-N" models mean only stage N selects 3$\times$3$\times$3 kernel, and others select 1$\times$5$\times$5. Note that we divide stage 4 of X3D with 11 layers into two stages (5 and 6 layers).}
\begin{floatrow}
% \scalebox{0.92}{
\resizebox{0.45\textwidth}{!}{
 \begin{tabular}{cccccc}
    	    \toprule
    		 Model &  Top1  &  \begin{tabular}[c]{@{}c@{}}Params\\ (M)\end{tabular} & \begin{tabular}[c]{@{}c@{}}FLOPs\\ (G)\end{tabular} & \begin{tabular}[c]{@{}c@{}}HomoEntr\\ Score\end{tabular}\\
    		\midrule
		  %  None & 44.4\% & 74.3\% & 181.55\\
		  %  S-2  &  \textbf{45.15\%} & \textbf{75.51\%} & 3.33 & 1.93\\
		  %  S-3  & 44.87\% & 75.02\% & 3.33 & 1.93\\
		  %  S-4  & 44.35\% & 74.69\% & 3.33 & 1.94 \\
		  %  S-5  & 43.85\% & 74.03\% & 3.33 & 1.94\\
		  %  S-6  & 43.83\% & 74.52\% & 3.33 & 1.94\\
		    S-2  &  \textbf{45.15\%} &  3.33 & 1.93 & 178.5\\
		    S-3  & 44.87\%  & 3.33 & 1.93 & 178.4\\
		    S-4  & 44.35\%  & 3.33 & 1.94 & 178.4\\
		    S-5  & 43.85\%  & 3.33 & 1.94 &178.4\\
		    S-6  & 43.83\%  & 3.33 & 1.94 &178.4\\
    		\bottomrule
    		
        \end{tabular}
}
% \scalebox{0.92}{
\resizebox{0.45\textwidth}{!}{
 \begin{tabular}{ccccc}
    	    \toprule
    		 Model &  Top1  &  \begin{tabular}[c]{@{}c@{}}Params\\ (M)\end{tabular} & \begin{tabular}[c]{@{}c@{}}FLOPs\\ (G)\end{tabular} & \begin{tabular}[c]{@{}c@{}}HomoEntr\\ Score\end{tabular}\\
    		\midrule
    % 		T-2  & 41.59\% & 70.51\% & 3.32 & 1.93\\
    % 		T-3  & 42.93\% & 72.83\% & 3.32 & 1.93\\
    % 		T-4  & 43.17\% & 72.81\% & 3.32 & 1.92\\
    % 		T-5  & \textbf{43.43\%} & \textbf{72.90\%} & 3.32 & 1.92\\
    % 		T-6  & 43.35\% & 73.57\% & 3.32 & 1.92\\
    		T-2  & 41.59\%  & 3.32 & 1.93 & 177.7\\
    		T-3  & 42.93\%  & 3.32 & 1.93 & 177.8\\
    		T-4  & 43.17\%  & 3.32 & 1.92 & 177.8\\
    		T-5  & \textbf{43.43\%} &  3.32 & 1.92 & 177.9\\
    		T-6  & 43.35\%  & 3.32 & 1.92 & 177.9\\
    		\bottomrule
    		
        \end{tabular}
}
\end{floatrow}
\label{tab:kernel}
\end{table}

% \begin{figure}[h]
%     \centering
%     \includegraphics[width=0.95\linewidth]{iclr2023/pics/kernel.pdf}
%     \caption{Illustrations of two 3D CNN kernel learning process: 1$\times$k$\times$k and k$\times$k$\times$k}
%     \label{fig:kernel examples}
% \end{figure}

\noindent\textbf{Kernel Selection Observations}.
% To figure out the impact of kernel on the accuracy and entropy,
% We conduct an experiment to explore the impact of different 3D convolutional kernel sizes on performance, as shown in Table \ref{tab:kernel}.
We conduct an experiment to explore how different 3D convolutional kernel sizes at different stages (\ie layer blocks at different positions in the network) impact the performance, as shown in Table \ref{tab:kernel}.
% \textcolor{blue}{All models are X3D-S \citep{feichtenhofer2020x3d} styled but with different kernels in different positions, 1$\times$5$\times$5 and 3$\times$3$\times$3 are typical 3D convolutional kernels for learning spatial and temporal information, respectively.?need to modify to explain why we use 155 and 33 for analysis, why not others, refer to the description in Movinet?}
All models are based on X3D-S but with different kernels in different stages. We set 1$\times$5$\times$5 and 3$\times$3$\times$3 kernels at the different stages in the 3D CNNs, which are typical 3D convolutional kernels for learning spatio-temporal information. 
These two different  choices enable a layer to aggregate the visual information focusing on different spatial and temporal dimensions, 
% different dimensional representations, expanding 
with the receptive field of CNN in the most pertinent directions.
% \textcolor{blue}{Note that we divide stage 4 with 11 layer of X3D into two stages (5 and 6 layers), to make the number of layers closer in each stage.?moving to caption is better?}
% great explanation like this.
% According to results in Table \ref{tab:kernel}, we obtain the following observations:
In Table \ref{tab:kernel}, the performances of S-2 and S-3 models are higher than X3D-S with only 3$\times$3$\times$3 kernels (44.6\% in Table \ref{tb:ss}), and S-series outperform T-series, which show that kernel selection at different stages influences the performance significantly, and that different stages may prefer different kernel sizes, respectively. 
% On the other hand, Table \ref{tab:kernel} shows that 
Although the kernel selections (with different spatio-temporal dimensions) at different stages lead to different effects on performance, the corresponding 3D CNNs have similar HomoEntr-Score. 
% It shows HomoEntr-Score's limitation on modeling expressiveness for 3D kernels. 
%%%%%%%%%%%%%%%%%%%%%%%%%%%%%%%%%%%%%%%%%%%%%%%%%%%%%%%%%%%%%%%%%%%%%%%%
% According to the results in Table \ref{tab:kernel}, we observe that: 
% (1) different kernel selections at different stages have different effects on performance, but similar HomoEntr-Score,
% and (2) the results of S-series outperform T-series and the performance of S-2 and S-3 models are higher than X3D-S (44.6\% in Table \ref{tb:ss} and all kernels are 3$\times$3$\times$3).

\begin{figure}[h]
    \centering
    \includegraphics[width=0.9\linewidth]{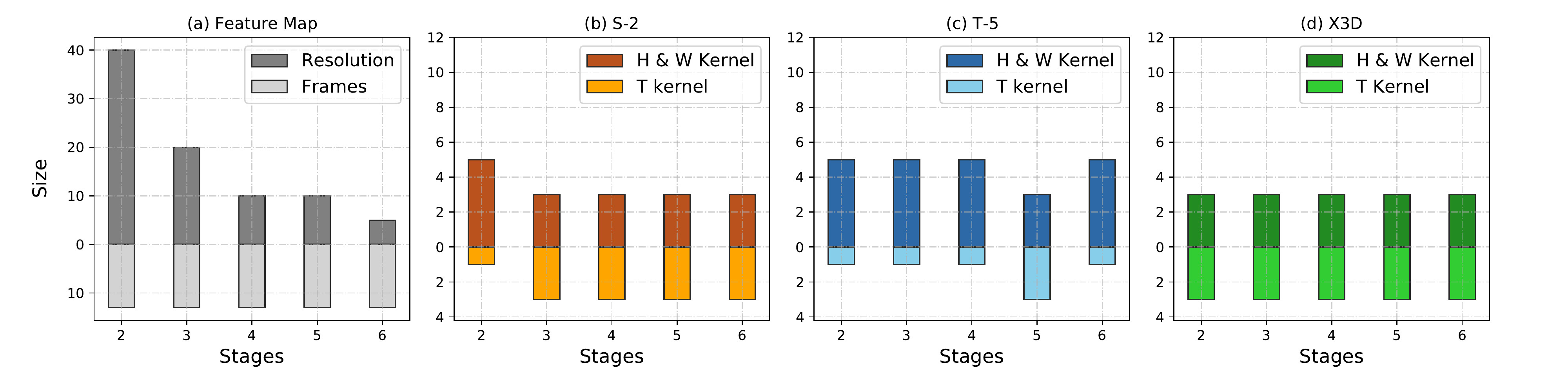}
    % \fbox{\rule{0pt}{1.5in} \rule{0.9\linewidth}{0pt}}
    \caption{Input feature map size and kernel sizes of S-2, T-5 and X3D model in each stage.}
    \label{fig:resolution}
\end{figure}

According to the downsampling strategy of the 3D CNNs, spatial resolutions become smaller from the large input as the depth increases, while the temporal frame size remains a certain value, as shown in Figure \ref{fig:resolution}a.
Through analyzing the results in Table \ref{tab:kernel}, we can infer that spatial kernels (like 1$\times$5$\times$5) can obtain spatial information more effectively at low-level stages, and spatio-temporal kernels (like 3$\times$3$\times$3) are more stable to obtain spatio-temporal information at high-level stages. 
Meanwhile, the similarity between the input feature map and the kernel size of the S-2 model at each stage is higher than that of the T-5 model or X3D, according to Figure \ref{fig:resolution}.
We thus consider that with the higher correlation between the feature map size and kernel size depth-wisely, the model can obtain higher expressiveness of spatial and temporal information.

\noindent\textbf{Spatio-Temporal Refinement}.
To estimate the correlation between feature map and kernel size in different depths, we first define two vectors: the input feature map size $\boldsymbol{S} = [T, H, W]$ and the 3D kernel size $\boldsymbol{K} = [K_t, K_h, K_w]$ in a convolutional layer, where $\{T, H, W\} \in \mathbb{R}$ represent frame, height and width dimension size.
% As cosine distance reflects the relative difference in vector direction and usually applied in the high-dimentional situation. 
% Thus, the cosine distance of $\boldsymbol{S}$ and $\boldsymbol{K}$ defined as:
We compute the distance $\hat{\mathcal{D}}$ based on commonly used cosine distance as:
% During the training process of 3D CNNs, the feature resolution will become smaller as the depth increases, but the frame number is fixed.
% Thus, to estimate the aggregation between temporal information and spatial information in different depths, we apply the cosine similarity and compute
% the \textit{Cosine Distance} between the input feature map size $\boldsymbol{S} = [T, H, W]$ and the 3D kernel size $\boldsymbol{K} = [K_t, K_h, K_w]$ in a convolutional layer, where $\{T, H, W\} \in \mathbb{R}$ represent frame, height and width dimension size.
% \begin{equation}
%     \mathcal{D}_{cosine}(X,K) = 1 - \frac{\Vert X\Vert_2\Vert K\Vert_2 - X\cdot K}{\Vert X\Vert_2\Vert K\Vert_2}
% \end{equation}
\begin{equation}
    \hat{\mathcal{D}}(\boldsymbol{S},\boldsymbol{K}) = -log(\mathcal{D}_{cosine}(\boldsymbol{S},\boldsymbol{K})) = -log\big(1 - \frac{\boldsymbol{S}\cdot \boldsymbol{K}}{\Vert\boldsymbol{S}\Vert \Vert\boldsymbol{K}\Vert}\big)~,
\end{equation}
where $\mathcal{D}_{cosine}$ represents the \textit{Cosine Distance} function, and we expand the diversity of the cosine distance by using $log$.
% W is not suitable, which will lead to a conflict with parameters. To unit with the Maximum Entropy Theory, I suggest $\alpha$ is more suitable to express the initialization of Weights.
% \textcolor{red}{As the initial variance of weight can be used to learn XXX ？dynamic}, 
We thus utilize the distance $\hat{\mathcal{D}}$ between $\boldsymbol{S}$ and $\boldsymbol{K}$ in each layer to define the variance of weight dynamically.
% \begin{equation}
%     \mathbb{D}^2({\boldsymbol{W}^l_{chw}}) = \hat{\mathcal{D}}(\boldsymbol{S}^l,\boldsymbol{K}^l) = -log[\mathcal{D}_{cosine}(\boldsymbol{S}^l,\boldsymbol{K}^l)]~.
% \end{equation}
Finally, we refine the upper bound differential entropy as:
\begin{equation}
    \mathcal{H}(F) \propto \sum^{L}_{l=1}log({{K^{l}_t}{K^{l}_h}{K^{l}_w}C^{l-1}}\cdot\hat{\mathcal{D}}(\boldsymbol{S}^{l},\boldsymbol{K}^l))~.
    \label{eq:st proxy}
\end{equation}
% \newpage
\begin{wrapfigure}{r}{6.2cm}
\captionsetup{font={small}}
    \centering
    \includegraphics[width=0.95\linewidth]{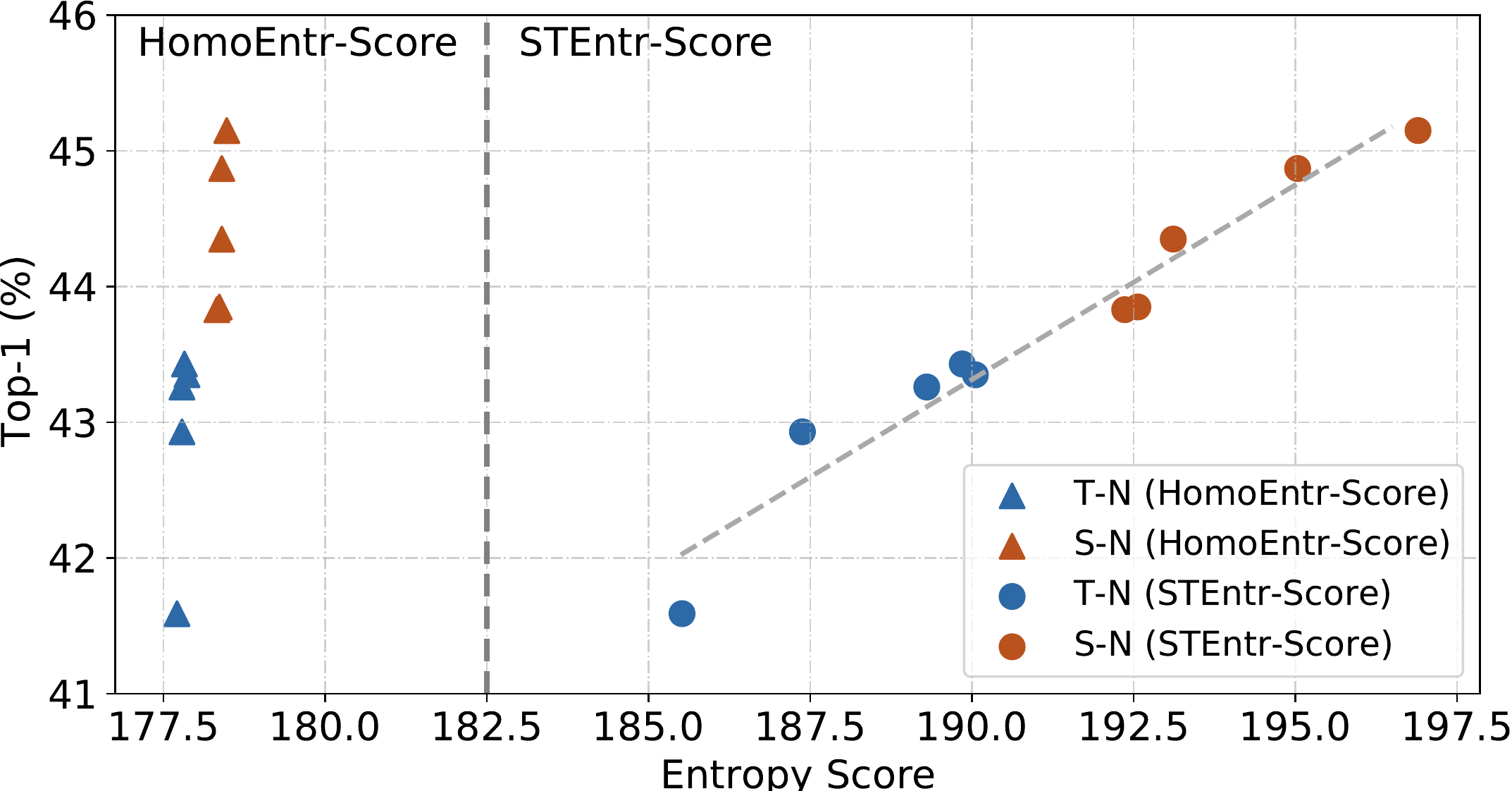}
    % \caption{Top-1 accuracy vs. STEntr-Score and HomoEntr-Score. `T-N' and `S-N' refer models in Table \ref{tab:kernel}.}
    \caption{Top-1 accuracy vs. STEntr-Score and HomoEntr-Score.}
    \label{fig:after calibration}
    % \vspace{-0.5cm}
\end{wrapfigure}
We name this analytic formulation of Eq. (\ref{eq:st proxy}) as \textbf{Spatio-Temporal Entropy Score} (STEntr-Score) to measure the aggregation of spatio-temporal dimensions.
After spatio-temporal refinement, we re-calculate the entropy of each model by STEntr-Score in Table \ref{tab:kernel}, and present the relationship between accuracy with STEntr-Score and HomoEntr-Score in Figure \ref{fig:after calibration}.
According to this figure, STEntr-Score is positively correlated with Top1 accuracy which indicates that the proposed spatio-temporal refinement can handle the discrepancy of visual information in spatial and temporal dimensions.
% \textcolor{blue}{STEntr-Score can thus help us rank various 3D CNNs architectures and evaluating without training during search}, and then \textcolor{red}{help} we can design a \textcolor{blue}{high-expressiveness} architecture for efficient action recognition.
% Therefore, STEntr-Score can be used to evaluate various 3D CNNs architectures without training during search.
% \textcolor{blue}{and help us design a high-expressiveness architecture for efficient action recognition.?Using this sentence to introduce the next section is better?}

% \begin{figure}[h]
%     \centering
%     \includegraphics[width=0.9\linewidth]{iclr2023/pics/ea.pdf}
%     % \fbox{\rule{0pt}{1.5in} \rule{0.9\linewidth}{0pt}}
%     \caption{Illustration of evolutionary algorithm applied in the E3D family architecture searching.}
%     \label{fig:ea}
% \end{figure}

\subsection{3D CNN Searching Strategy}
% \textcolor{blue}{The designed Spatio-Temporal Entropy--a proxy represents the information capacity of network-- builds a connection between 3D network architecture and training accuracy. Thus, the proxy should combine with a specific searching strategy with a search space, which means our entropy is not tightly coupled with searching strategies (e.g., Reinforcement Learning, EA and so on.), enabling flexibility in the network design.}
% is indeed not 
Utilizing STEntr-Score, we apply the basic \textbf{Evolutionary Algorithm} (EA) to find the optimal 3D CNN architectures, which is similar to \citep{lin2021zen,sun2022mae}.
%  in the \textcolor{blue}{simple network space
% \textcolor{blue}{1. Block-wise searching. 2. Initialization model; 3.Kernel options. 4.Width: channels and bottleneck channels. 5. Block Layers}
We initialize a population of candidates randomly under a small budget and define the 3D kernel search space within each layer with two options: \{$1\times{(k^{space})^2}$, $k^{times} \times (k^{space})^2$\}, then randomly select two stages from the candidates and mutate them at each iteration step.
We calculate its STEntr-Score to navigate the evolution process instead of evaluating the accuracy after mutation, if the inference cost of the mutated structure does not exceed the budget.
The population will be maintained to a certain size during iterations, by discarding the worst candidate of the smallest STEntr-Score.
% \textcolor{blue}{After mutation, if the inference cost of mutated structure does not exceed the budget, we calculate its STEntr-Score which helps to navigate the evolution process instead of accuracy.
% During iterations, the population is maintained to a certain size by discarding the worst candidate of the smallest STEntr-Score.}
% \textcolor{blue}{next calculating STEntr-Score for rank.
% Throughout the search process, STEntr-Score is applied to navigate the evolution process instead of accuracy, which means we do not train on the dataset. During EA iterations,
% the population is maintained to a certain size by discarding the worst candidate of the smallest STEntr-Score.}
After all iterations, the target network is achieved with the largest STEntr-Score under the given budget (\eg FLOPs, parameters, and latency).
% \textcolor{blue}{Move to experiments: Thus, we obtain the spatio-temporal entropy 3D CNNs family, consists of E3D-S, E3D-M, and E3D-L.}
% Note that, we only apply the STEntr-Score to guide the evolution process, not accuracy, which does not need training on the dataset in the calculation process of STEntr-Score.
% \textcolor{red}{It is possible to choose other methods such as reinforcement learning or even greedy selection. ? The choice of EA is due to its simplicity ? I think this explanation is not suitable here.
Since the latency budget requires a forward process on GPU which will diminish the efficiency of our STEntr-Score search, we choose FLOPs as the target budget. Another reason for applying FLOPs budget is to fairly compare with X3D \citep{feichtenhofer2020x3d} and MoViNet \citep{kondratyuk2021movinets}, which only report FLOPs rather than latency.
% , and latency budget requires forward network on GPU which is not efficiency during searching process. 
Thus, we obtain the spatio-temporal entropy 3D CNNs family (E3D family) under certain FLOPs,
All models are searched separately with different FLOPs bugdet (1.9G, 4.7G, and 18.4G) for a fair comparison with X3D-S/M/L as the baseline, and the detailed algorithm is described in \textbf{Appendix \ref{sec:algorithm}}
% The detailed settings are provided in Section 4.1 and detailed algorithm is described in \textbf{Appendix \ref{sec:algorithm}}.
% and detailed structure of E3D family is described in \textbf{Appendix \ref{sec:architecture}}

\section{Experiments}
% Based on evolutionary algorithm, 
Our E3D family consists of E3D-S (1.9G FLOPs), E3D-M (4.7G FLOPs), and E3D-L (18.3G FLOPs).
The detailed structures of the E3D family are described in \textbf{Appendix \ref{sec:architecture}}.
% We present our experimental results on three large-scale public datasets.
We compare our approach with other state-of-the-art methods and in-depth analysis to better understand our method.
More results are presented in \textbf{Appendix \ref{sec:more results}}.

% \newpage
\subsection{Experiment Settings}
The E3D family includes a search stage without training and a training \& inference stage for video recognition on a specific dataset. Detailed settings in each stage are described as follows:
% \noindent\textbf{Datasets}.
% Our experiments are conducted on three large-scale datasets, Something-Something (Sth-Sth) V1\&V2 \cite{goyal2017something}, and Kinetics400 \cite{kay2017kinetics}. More dataset details can be seen in the supplementary materials.
% 1) The Sth-Sth datasets are more focused on fine-grained and motion-dominated actions, which contain pre-defined basic actions involving different interacting objects. 
% Sth-Sth V1 comprises 86k video clips in the training set and 12k video clips in the validation set. Sth-Sth V2 is an updated version of Sth-Sth V1, which contains 169k video clips in the training set and 25k video clips in the validation set. They both have 174 action categories.
% 2) The Kinetics dataset contains activities in daily life and some categories are highly correlated with interacting objects or scene context.
% Kinetics400 contains over 200k training videos and 20k validation videos divided into 400 categories, covering a wide range of human activities.

% \subsection{Implementation Details}

\noindent\textbf{Search Settings}. 
Following X3D \citep{feichtenhofer2020x3d}, we also apply a MobileNet-like network basis, in which the core concept is 3D depth-wise separable convolution for efficiency.
The initial structure is composed of 5 stages with small and narrow blocks to meet the reasonable budget, which is usually below 1/3 of the target FLOPs budget.
The population size and total iterations of EA are set as 512 and $500000$, respectively.
% with layer arrangement (1-\textcolor{red}{85}).
During mutation stages from the candidates,
we randomly select 3D kernels from \{1$\times$3$\times$3, 1$\times$5$\times$5, 3$\times$3$\times$3\} to replace the current one; interchange the expansion ratio of bottleneck from $\{1.5, 2.0, 2.5, 3.0, 3.5, 4.0\}$($bottleneck = ratio\times intput$); scale the output channels with the ratios $\{2.0, 1.5, 1.25, 0.8, 0.6, 0.5\}$; or increases or decreases depth with 1 or 2. 
% These mutation strategies will assure the diversity of variants. 
Note that the channel dimension of every layer is fixed within 8 to 640 with multiples of 8, which helps shrink homologous search space and accelerate search speed.
% Some kernel sizes may benefit from having different numbers of input filters, so we search over a range of expanding bottleneck widths in and output channels with 

\noindent\textbf{Training \& Inference}.
Our experiments are conducted on three large-scale datasets, Something-Something (Sth-Sth) V1\&V2 \citep{goyal2017something}, and Kinetics400 \citep{kay2017kinetics}.
All models are trained by using Stochastic Gradient Descent (SGD).
The cosine learning rate schedule \citep{loshchilov2016sgdr} is employed, and total epochs are set to 100 for Sth-Sth V1\&V2 datasets, and 150 for Kinetics400 dataset, with synchronized Batch-Norm instead of common Batch-Norm. 
% To make the training robust, a warm-up strategy is also conducted.
Random scaling, cropping, and flipping are applied as data augmentation on all datasets.
% Besides, all experiments are performed on 8$\times$Nvidia Tesla A100 GPUs.
% \noindent\textbf{Inference}.
To be comparable with previous work and evaluate accuracy and complexity trade-offs, we apply two testing strategies:
1) \textit{K-Center}: temporally, uniformly sampling of K clips (\eg K=10) from a video and taking a center crop.
2) \textit{K-LeftCenterRight}: also uniformly sampling K clips temporally, but taking multiple crops to cover the longer spatial axis, as an approximation of fully-convolutional testing.
For all methods, we follow prior studies by reporting Top1 and Top5 recognition accuracy, and FLOPs to indicate the model complexity.
More experiment setting details can be seen in \textbf{Appendix \ref{sec:setting}}.

\begin{table}[]
\captionsetup{font={small}}
\caption{Comparison with state-of-the-art methods on Sth-Sth V1 and V2 validation datasets. The models only take RGB frames as inputs. To be consistent with compared approaches, we present most results of 2D CNN-based methods with ResNet50. ``MN-V2"" denotes MobileNet-V2. k$\times$k denotes temporal clip with spatial crop evaluation. “-” indicates the results are not available for us, and $*$ denotes our reproduced models.}
\label{tb:fps}
    \centering
    % \footnotesize
     \resizebox{\textwidth}{!}{
        \begin{tabular}{l  c  c  c  c  c  c  c  c}
        \toprule
        \multirow{2}{*}{Method} &
        \multirow{2}{*}{Backbone} &
        \multirow{2}{*}{Pretrain} &
        \multirow{2}{*}{Resolution} &
        \multirow{2}{*}{GFLOPs} &
        \multicolumn{2}{c}{1$\times$1 V1-Val}    &
        % \multicolumn{2}{c}{1$\times$1 V2-Val}    &
        \multicolumn{2}{c}{2$\times$3 V2-Val}     \\
        \cmidrule(lrr){6-7}
        \cmidrule(lrr){8-9}
        % \cmidrule(lrr){10-11}
        
        \multicolumn{1}{c}{} &
        \multicolumn{1}{c}{} &
        \multicolumn{1}{c}{} &
        \multicolumn{1}{c}{} &
        \multicolumn{1}{c}{} &
        \multicolumn{1}{c}{Top1} &
        \multicolumn{1}{c}{Top5}   &
        % \multicolumn{1}{c}{Top1} &
        % \multicolumn{1}{c}{Top5}   &
        \multicolumn{1}{c}{Top1}   & 
        \multicolumn{1}{c}{Top5}    \\ 
        % Method & Pretrain &  Frame & Resolution & \#Param. & GFLOPs & Top-1 & Top5\\ 
        \midrule
        TSN \citep{wang2016tsn} & ResNet50 & ImageNet & 8 $\times$ 256$^2$ & 16  & 19.5 & -  & - &  -\\
        % TRN-Multiscale\citep{zhou2018trn} & BNInception & ImageNet & 8  & 33  & 34.4 & -  & - & -\\
        % ECO \citep{zolfaghari2018eco} & BNIncep+Res18 & K400 & 16  & 64  & 41.6 & - & - & -\\
        % TSM \citep{lin2019tsm} & ResNet50 & ImageNet & 8 $\times$ 256$^2$ & 33  & 45.6 & 74.2  & 63.4 & 88.5\\
        TSM \citep{lin2019tsm} & ResNet50 & ImageNet & 16 $\times$ 256$^2$ & 65  & 47.2 & 77.1  &  63.4 & 88.5\\
        % TIN \citep{shao2020tin} & ResNet50 & ImageNet & 8  & 34  & 45.8 & 75.1  & 60.0 & 85.5 & - & -\\
        % TIN \citep{shao2020tin} & ResNet50 & ImageNet & 16  & 67  & 47.0 & 76.5  & - & - & - & -\\
        % TEINet \citep{liu2020teinet} & ResNet50 & ImageNet & 8 $\times$ 256$^2$ & 33  & 47.4 & -  & - & -\\
        % bLVNet-TAM \citep{liu2020teinet} & bLResNet-50 & ImageNet & 8$\times$2  & 23.8  & 46.4 & 76.6  & 59.1 & 86.0 & - & -\\
        % TANet \citep{liu2021tam} & ResNet50 & ImageNet & 8 $\times$ 256$^2$ & 33 & 47.3 & 75.8  & 62.7 & 88.0\\

        TANet \citep{liu2021tam} & ResNet50 & ImageNet & 16 $\times$ 256$^2$ & 66 & 47.6 & 77.7 & 64.6 & 89.5\\
        % TEA \citep{li2020tea} & ResNet50 & ImageNet & 8 $\times$ 256$^2$  & 66 & 48.9 & 78.1  & - & -\\
        ActionNet \citep{wang2021actionnet} & ResNet50 & ImageNet & 16 $\times$ 256$^2$ & 69.5  & - & -  & 64.0 & 89.3\\
        TAda \citep{huang2021tada} & ConvNeXt-T & ImageNet & 16 $\times$ 256$^2$ & 47 & - & - & 64.8 & 88.8 \\
        % \textcolor{blue}{AdaFocusV2-TSM \citep{wang2022adafocus}} & \textcolor{blue}{MN-V2+ResNet50} & \textcolor{blue}{ImageNet} & \textcolor{blue}{(8+12) $\times$ 176$^2$} & \textcolor{blue}{33.7} & \textcolor{blue}{49.6} & - & \textcolor{blue}{61.3} & - \\
        \midrule
        I3D \citep{carreira2017i3d}  & InceptionV1 & ImageNet+K400 & 64 $\times$ 256$^2$ & 306  & 41.6 & 72.2  & - & -\\
        NL I3D \citep{carreira2017i3d}  & InceptionV1 & ImageNet+K400 & 64 $\times$ 256$^2$ & 334  & 44.4 & 76.0 & - & -\\
        % RubiksNet-L \citep{fan2020rubiksnet}  & - & No pretrain & -  & 15.8  & - & -  & 61.7& 87.3\\
        S3D-G \citep{xie2018s3d} & InceptionV1 & ImageNet & 64 $\times$ 256$^2$ & 71.4  & 48.2 & 78.7  & - & -\\
        X3D$*$ \citep{feichtenhofer2020x3d} & X3D-S & No pretrain & 13 $\times$ 160$^2$ & 2  & 44.6 & 74.4   & 60.1 & 85.9\\
        X3D$*$ \citep{feichtenhofer2020x3d} & X3D-M & No pretrain & 16 $\times$ 224$^2$ & 4.7  & 47.3 & 76.6   & 62.2 & 87.2\\
        X3D$*$ \citep{feichtenhofer2020x3d} & X3D-L & No pretrain & 16 $\times$ 312$^2$ & 18.4  & 49.4 & 77.9   &  & \\
        MoViNet$*$ \citep{kondratyuk2021movinets} & MoViNet-A0 & No pretrain & 50 $\times$ 172$^2$ & 2.7  & 46.9 & 75.0  & 61.9 & 87.2\\
        MoViNet$*$ \citep{kondratyuk2021movinets} & MoViNet-A1 & No pretrain & 50 $\times$ 172$^2$ & 6  & 49.3 & 77.1  & 64.5 & 89.1\\
        % MoViNet-A1$*$ \citep{kondratyuk2021movinets} & 3D & No pretrain & 13 $\times$ 256$^2$ & 1.1  & 44.4 & 74.5  & - & -\\
        \midrule
        E3D & E3D-S & No pretrain & 13 $\times$ 160$^2$  & 1.9 & 47.1 & 75.6   & 62.1 & 87.6 \\
        E3D & E3D-M & No pretrain & 16 $\times$ 224$^2$  & 4.7 & 49.4 & 78.1  & 64.7 & 89.6 \\
        E3D & E3D-L & No pretrain & 16 $\times$ 312$^2$  & 18.3 & \textbf{51.1} & \textbf{78.7}  & \textbf{65.7} & \textbf{89.8} \\
        \bottomrule
        \end{tabular}
        }
        \label{tb:ss}
\end{table}

\begin{table}[]
\captionsetup{font={small}}
\caption{Comparison with state-of-the-art methods on the validation set of Kinetics400. We report the inference cost with a single “view" (temporal clip with spatial crop) × the number of such views used (GFLOPs$\times$views). “N/A” and “-” indicate the numbers are not available for us.}
\label{tb:fps}
    \centering
    % \footnotesize
     \resizebox{\textwidth}{!}{
        \begin{tabular}{l c c  c  c  c  c  c c}
        \toprule[1pt]
        \multirow{2}{*}{Method} &
        \multirow{2}{*}{Backbone} &
        \multirow{2}{*}{Pretrain} &
        \multirow{2}{*}{Frame} &
        % \multirow{2}{*}{Resolution} &
        \multirow{2}{*}{\#Param.} &
        \multirow{2}{*}{GFLOPs $\times$ Views} &
        \multicolumn{2}{c}{Val}    \\
        \cmidrule(lrr){7-8}
        
        \multicolumn{1}{c}{} &
        \multicolumn{1}{c}{} &
        \multicolumn{1}{c}{} &
        \multicolumn{1}{c}{} &
        \multicolumn{1}{c}{} &
        \multicolumn{1}{c}{} &
        \multicolumn{1}{c}{Top1}   & 
        \multicolumn{1}{c}{Top5}    \\ 
        % Method & Pretrain &  Frame & Resolution & \#Param. & GFLOPs & Top-1 & Top5\\ 
        \midrule[1pt]
        % MF-Net \citep{chen2018mfnet} & ResNet50 & ImageNet & 16  & 8M & 11.1$\times$1$\times$50 & 72.8\% & 90.4\%  \\ 
        % ip-CSN \citep{tran2019csp} & ResNet50 & ImageNet & 8  & - & 1.2$\times$10$\times$1 & 70.8\% & -  \\
        TSN \citep{wang2016tsn} & ResNet50 & ImageNet & 25  & 24.3M & 80$\times$1$\times$10 & 72.5 & 90.2  \\
        % STM \citep{jiang2019stm} & ResNet50 & ImageNet & 16  & - & 67$\times$3$\times$10 & 73.7\% & 91.6\%  \\
        TSM \citep{lin2019tsm} & ResNet50 & ImageNet & 16  & 24.3M & 65$\times$3$\times$10 & 74.7 & 91.4  \\
        % TEA \citep{li2020tea}& ResNet50 & ImageNet & 8  & - & 35$\times$3$\times$10 & 75.0\% & 91.8\%  \\
        TEA \citep{li2020tea}& ResNet50 & ImageNet & 16  & - & 70$\times$3$\times$10 & 76.1 & 92.5  \\
        % TEINet \citep{liu2020teinet} & ResNet50 & ImageNet & 16  & 25.1M & 86$\times$3$\times$10 & 76.2\% & 92.5\%  \\
        % TANet \citep{liu2021tam} & ResNet50 & ImageNet & 8  & 26M & 43$\times$3$\times$10 & 76.3\% & 92.6\%  \\
        TANet \citep{liu2021tam} & ResNet50 & ImageNet & 16  & 26M & 86$\times$3$\times$10 & 76.9 & 92.9  \\
        TDN \citep{wang2021tdn} & ResNet50 & ImageNet & 16+64  & - & 72$\times$3$\times$10 & 77.5 & 93.2  \\
        % TAda2D \citep{huang2021tada} & ResNet50 & ImageNet & 16  & 26M & 86.2$\times$3$\times$10 & 77.4\% & 93.1\%  \\
        R(2+1)D \citep{tran2018R21d} & ResNet50 & ImageNet & 16  & 63.6M & 67$\times$3$\times$10 & 73.7 & 91.6  \\
        % bLVNet-TAM \citep{liu2020teinet} & bLResNet-50 & ImageNet & 48  & - & 93$\times$3$\times$3 & 73.5\% & 91.2\%\\
        SlowOnly \citep{feichtenhofer2019slowfast} & ResNet50 & ImageNet & 8  & - & 42$\times$3$\times$10 & 74.8 & 91.6\\
        SlowFast \citep{feichtenhofer2019slowfast} & ResNet50 & ImageNet & 8+32  & 34.4M & 65.7$\times$3$\times$10 & 77.0 & 92.6\\
        % SlowFast \citep{feichtenhofer2019slowfast} & ResNet50 & ImageNet & 8+32  & 34.4M & 66$\times$3$\times$10 & 77.0\% & 92.6\%\\
        \midrule[1pt]
        I3D \citep{carreira2017i3d} & InceptionV1 & ImageNet & 64  & 12M & 108$\times$N/A & 72.1 & 90.3  \\ 
        Two-Stream I3D \citep{carreira2017i3d} & InceptionV1 & ImageNet & 64  & 25M & 216$\times$N/A & 75.7 & 92.0  \\
        % NL I3D \citep{carreira2017i3d} & InceptionV1 & ImageNet & 128  & 12M & 282$\times$3$\times$10 & 76.5\% & 92.6\%  \\ 
        S3D-G \citep{xie2018s3d} & InceptionV1 & ImageNet & 64  & - & 71.4$\times$3$\times$10 & 74.7 & 93.4  \\ 
        X3D \citep{feichtenhofer2020x3d} & X3D-M & No pretrain & 16  & 3.8M & 6.2$\times$3$\times$10 & 76.0 & 92.3  \\
        X3D \citep{feichtenhofer2020x3d} & X3D-L & No pretrain & 16  & 3.8M & 24.8$\times$3$\times$10 & 77.5 & 92.9  \\
        MoViNet \citep{kondratyuk2021movinets} & MoViNet-A2 & No pretrain & 50 & 4.6M & 10.3$\times$1$\times$1 & 75.0 & 92.3 \\
        % MoViNet \citep{kondratyuk2021movinets} & MoViNet-A3 & No pretrain & 50 & 5.4M & 56.9$\times$1$\times$1 & 78.2 & 93.8 \\
        \midrule[1pt]
        TimeSformer-S \citep{bertasius2021timesformer} & ViT-B & ImageNet & 8  & 121.4M & 590$\times$3$\times$10 & 78.0 & 93.7  \\
        % ViT-B-VTN \citep{neimark2021video} & ViT-B & ImageNet & 8  & 110M & 4218$\times$1$\times$1 & 78.6 & 93.7  \\
        Swin \citep{liu2022video} & Swin-T & ImageNet & 32 & 28.2M & 88$\times$3$\times$4 & 78.8 & 93.6 \\
        \midrule[1pt]
        % E3D & E3D-S & No pretrain & 16  & 3.4M & 1.9$\times$3$\times$10 & 74.8 & 91.4 \\
        E3D & E3D-M & No pretrain & 16  & 3.4M & 4.7$\times$3$\times$10 & 76.4 & 92.5 \\
        E3D & E3D-L & No pretrain & 16  & 5.8M & 18.3$\times$3$\times$10 & 77.6 & 92.9 \\
        \bottomrule[1pt]
        \end{tabular}
        }
        \label{tb:k400}
\end{table}

\subsection{Main Results}

\noindent\textbf{Sth-Sth V1\&V2}.
Tabel \ref{tb:ss} shows the comparison between E3D family and state-of-the-art methods.
It can be seen that our proposed E3D family achieves competitive performance with more efficient FLOPs-level, which indicates that the E3D models can recognize actions effectively and efficiently.
1) Compared to 2D CNN-based methods, E3D outperforms most previous approaches on the same FLOPs-level.
% Note that 2D CNN-based methods do not provide results of mobile-level backbone, and the FLOPs of these methods is significant higher than ours.
Even compared to many methods with similar performance, our model requires much lower computational costs.
Note that our E3D family does not need to be pretrained on other datasets, and the performance of these 2D CNN-based methods is based on ResNet50 or a stronger backbone that is not suitable for low-level computation.
2) The E3D family also achieves higher performance compared to 3D CNN-based methods, which indicates that the architecture of E3D can      handle the discrepancy of visual information in spatial and temporal dimensions
Compared to the NAS-based method \citep{kondratyuk2021movinets}, our proposed E3D can still achieve a remarkable result which thus verifies the effectiveness of the STEntr-Score for searching the architecture.

\noindent\textbf{Kinetics400}.
% We also conduct comparison with state-of-the-art methods on Kinetics400 dataset.
Table \ref{tb:k400} shows that E3D achieves state-of-the-art performance compared to most 2D and 3D methods, but uses much less computational resources.
1) Most methods apply ImageNet pretrained backbones on the Kinetics400 dataset. However, our model can still achieve excellent results without using pretrained models, which indicates that our searched architecture by STEntr-Score can effectively learn spatio-temporal information.
2) E3D outperforms other 3D CNN-based models \citep{carreira2017i3d,xie2018s3d,feichtenhofer2020x3d} which only employ 3$\times$3$\times$3 kernel. It means that kernel selection is important for action recognition, and STEntr-Score can benefit 3D CNN architecture design. 
3) Even though the performance of Transformer-based models \citep{bertasius2021timesformer,neimark2021video,liu2022video} is competitive, our model still provides remarkable results by using much lower computational resources (FLOPs) and parameters, which means our model is more suitable in efficient scenarios.

\subsection{Correlation Study}

\begin{figure}[h]
    \centering
    \includegraphics[width=0.9\linewidth]{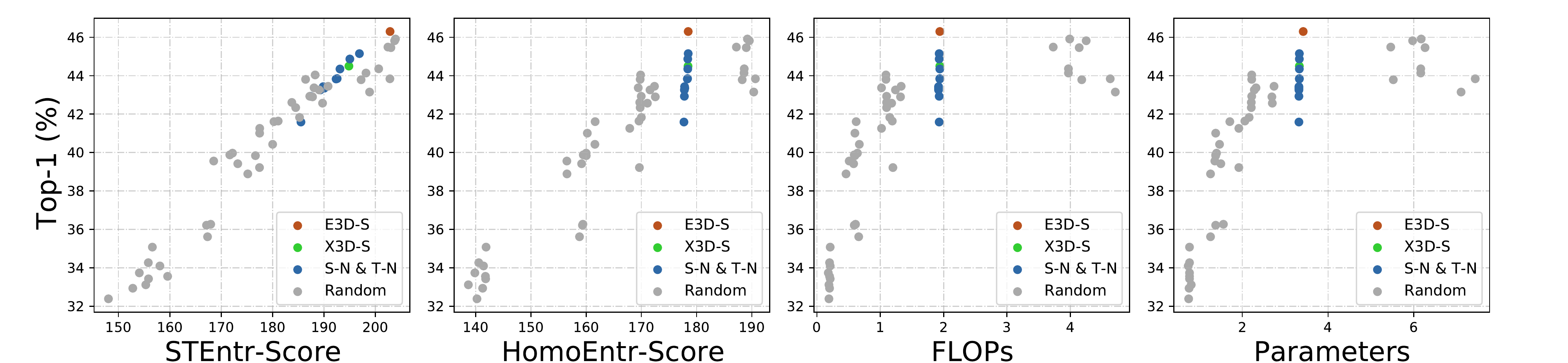}
    \caption{Correlations between Top1 and STEntr-Score, HomoEntr-Score, FLOPs, and Parameters. Points represent different sampled models, which have different channel numbers and layer configurations.}
    \label{fig:corre}
\end{figure}

To verify the importance of STEntr-Score in the design of video understanding models, we randomly construct 60 different models (0.2 to 5 GFLOPs) with different channel dimensions and layer numbers to investigate the correlations between STEntr-Score, HomoEntr-Score, FLOPs and parameters. For a fair comparison, all networks are trained on the Sth-Sth V1 dataset with batch size of 256 and 50 epochs.
We also provide the performance of E3D-S and X3D-S under the same training setting. According to results in Figure~\ref{fig:corre}, we can observe that:
(1) The proposed STEntr-Score is more positively correlated with Top1 accuracy than other metrics, which proves the effectiveness of our proposed STEntr-Score in evaluating network architecture.
(2) Although HomoEntr-Score is discriminative on different FLOPs levels, the ability to capture the discrepancy of the visual information in the spatial and temporal domain is not as good as STEntr-Score on the same FLOPs level.
(3) Benefiting from STEntr-Score, EA can help us obtain 3D CNN architectures with higher expressiveness as measured by STEntr-Score on the same FLOPs or parameters level.

% The comparison between STEntr-Score and HomoEntr-Score indicates that our proposed method can successfully measure the spatio-temporal aggregation of 3D CNNs architecture (\textcolor{red}{standard can measure but not st}).  

% wisely

\subsection{Discussion}
\begin{figure}[h]
	\centering
    \begin{subfigure}[b]{0.47\textwidth}
  \centering
  \includegraphics[width=1\textwidth]{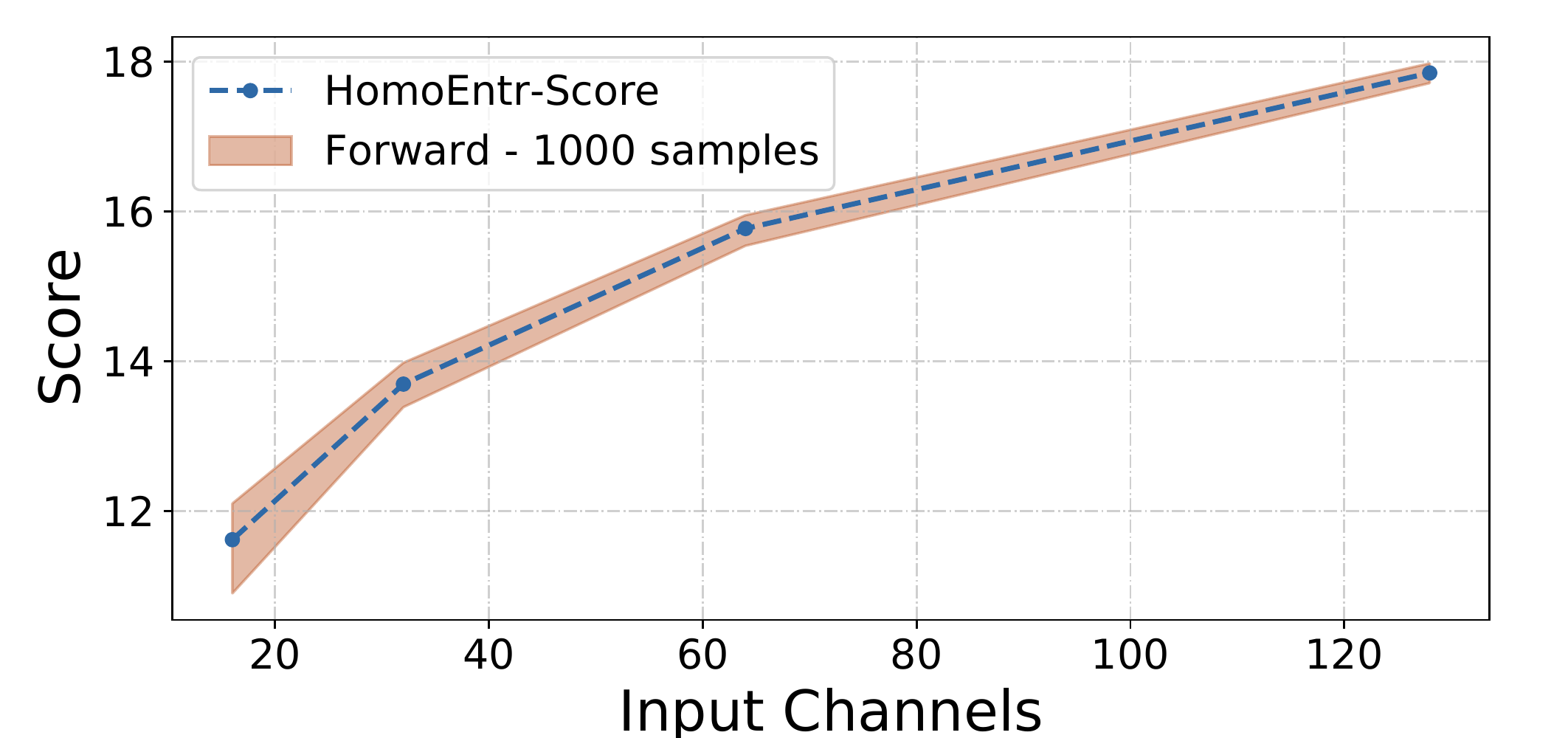}
%   \caption{Score vs.. Channels.}
  \caption{Consistency comparison.}
  \label{sfig:score}
    \end{subfigure}
  \begin{subfigure}[b]{0.47\textwidth}
  \centering
  \includegraphics[width=1\textwidth]{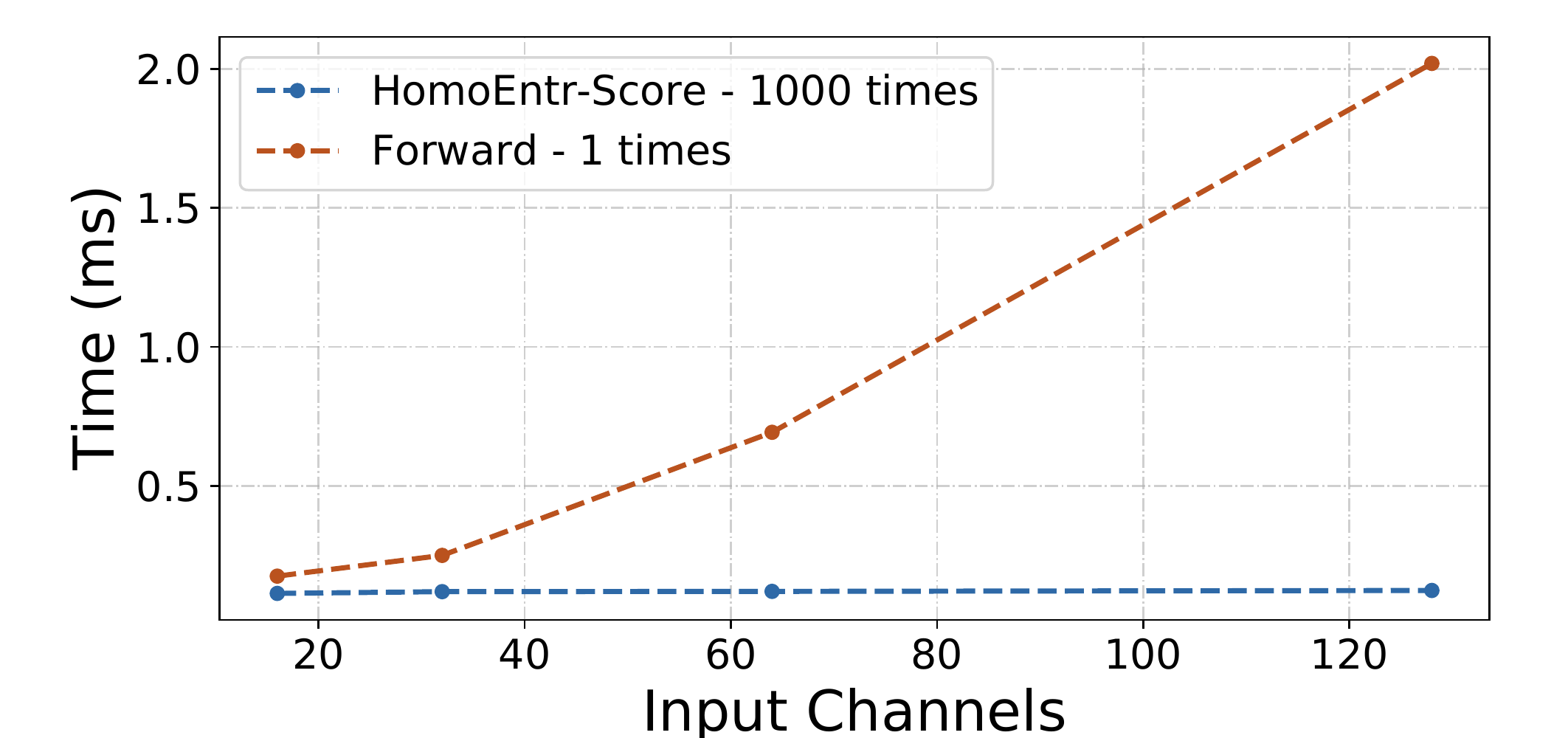}
%   \caption{Time vs.. Channels.}
  \caption{Efficiency comparison.}
  \label{sfig:time}
    \end{subfigure}
	\caption{Comparisons between HomoEntr-Score and “Forward" calculations. “Forward" represents using the forward inference. The calculations are conducted on an AMD Ryzen 5 5600X 6-core CPU.}
	\label{fig:simulation}	
\end{figure}

\noindent\textbf{Comparison with forward inference}.
% \textcolor{blue}{To this end, we have derive the mathematical expression of the entropy, which have several advantages over the forward entropy~\citep{sun2022mae}. }
% To verify the stability and speed of computing the MaxE score, 
For a fair comparison with the realization of forward inference in~\citep{sun2022mae}, we use HomoEntr-Score and conduct a simulation in a three-layer 3D network.
The shape of the input feature is $5\times5\times5$, kernel sizes are set to $1\times1\times1$, $3\times3\times3$ and $1\times1\times1$ with a stride of $1$, and channels are all set to $C_{in}\in\{16,32,64,128\}$. 
The entropy of each network is calculated $10^3$ times with either forward inference via Eq. (\ref{eq:gaussian}) or direct computation of HomoEntr-Score. 
When performing the forward inference of the network, convolution blocks are re-initialized based on a Gaussian distribution during each iteration.
% Meanwhile, the entropy is also calculated with Eq. \ref{eq:x_d} with $10^3$ times. 
% Results are presented in Figure~\ref{fig:simulation}.
The filled “Forward" range in Figure~\ref{sfig:score} demonstrates there exists variance between different random samples, which also emphasizes the stability of the analytic formulation. In Figure~\ref{sfig:time}, regardless of how channels change, the speed of $10^3$ times formulaic calculation of value remains constant, while the speed reduces almost linearly when performing forward inference. More comparison analysis of training-free scores is included in \textbf{Appendix \ref{sec:entropy comparison}}

% \noindent\textbf{Search Strategy Discussion}.

\begin{wraptable}{r}{7.1cm}
    \caption{Searching cost comparison on the Sth-Sth V1 dataset. $\S$: 64 Google TPUv3, Power 450W per TPUv3; $\dag$: 1 AMD Ryzen 5 5600X 6-Core CPU, Power 65W;}
    \centering
    \resizebox{\textwidth}{!}{
    \begin{tabular}{lccccc}
    	     \toprule
    		 Method & \begin{tabular}[c]{@{}c@{}}Search \\ Devices\end{tabular} & \begin{tabular}[c]{@{}c@{}}Search \\Time \end{tabular} & \begin{tabular}[c]{@{}c@{}}Power \\Consumption\end{tabular}& GFLOPs &  TOP-1\\
    		\midrule
    		MoViNet-A1 & TPUs $\S$ & 24h  & 691.2kWh & 6 & 49.3 \\
		    E3D-M & CPU $\dag$ & 3h  & 0.195kWh & 4.7 & 49.4 \\
    		\bottomrule
    		
        \end{tabular}}
    
    \label{tab:time}
\end{wraptable}
\noindent\textbf{Searching cost comparison}.
% \textcolor{blue}{?Since we have a emphasis on searching efficiency, please mention some advantage of MoViNet because this paper may be reviewed by them.?}
Since we apply analytic formulation rather than inference, the calculation of our STEntr-Score has lower hardware requirements, which means that CPU resources can meet it instead of GPU or TPU.
From Table~\ref{tab:time}, our method only takes three hours of searching time with a desktop CPU, while MoViNet consumes 24 hours with 64 commercial TPUs.
Extremely low time and power consumption demonstrate the searching efficiency of our analytic entropy formulation.
% However, the performance of our method remains similar which also verify the effectiveness of STEntr-Score.}

\section{Conclusion}

In this paper, we propose to automatically design efficient 3D CNN architectures via an entropy-based training-free neural architecture search approach, to address the problem of efficient action recognition.
In particular, we first formulate the 3D CNN architecture as an information system and propose the STEntr-Score to measure the expressiveness of the system.
Then we obtain the E3D family by an evolutionary algorithm, with the help of STEntr-Score.
Extensive results show that our searched E3D family achieves higher accuracy and better efficiency compared to many state-of-the-art action recognition models, within three desktop CPU hours searching.

\subsubsection*{Acknowledgments}
% This research is supported by Alibaba Group through Alibaba Research Intern Program, and D. Gong was partially supported by ARC DECRA Fellowship DE230101591.

This research was supported by Alibaba Group through Alibaba Research Intern Program, and ARC DECRA Fellowship DE230101591 to D. Gong.

\bibliography{maxe3d}
\bibliographystyle{iclr2023_conference}

\newpage
\appendix
\section*{Appendix}

% \tableofcontents

In the appendix, we provide a detailed description of notations in this paper (Appendix A), detailed proof of equations (Appendix B), a comparison between different training-free scores on the ImageNet dataset (Appendix C),  a discussion of simple network space in the entropy mechanism (Appendix D), STEntr-Score for maximizing expressiveness (Appendix E), E3D family structure details (Appendix F), experimental setting details (Appendix G), additional result analysis (Appendix H), and future work discussion (Appendix I).

\section{Meaning of Notations}
\label{sec:notation}

\begin{table}[h]
    \centering
    \caption{The meaning of all notations appeared in this paper.}
    \resizebox{\textwidth}{!}{
    \begin{tabular}{ccl}
    \toprule
       Notation  &  Size &  Meaning \\
    \midrule
        $\mathcal{H}(x)$ & - & The function of computing the entropy of the given $x$\\
        $P(x)$ & - & The distribution of the given $x$\\
        $\mu$ & Constant & The value of expectation\\
        $\mathbb{E}$ & - & The function of computing expectation\\
        $\sigma$ & Constant & The value of variance\\
        $\mathbb{D}$ & - & The function of computing variance\\
        $K_t$ & Constant & Temporal dimension size of 3D CNN kernel\\
        $K_h$ & Constant & Height dimension size of 3D CNN kernel\\
        $K_w$ & Constant & Width dimension size of 3D CNN kernel\\
        $C$ & Constant & Channel dimension size\\
        $\boldsymbol{K}$ & $K_t \times K_h \times K_w$ & A 3D CNN kernel size\\
        $\boldsymbol{W}$ & $C \times \boldsymbol{K}$ & The weight matrix of the CNN layer\\
        $\boldsymbol{S}$ & $T \times H \times W$ & The input feature map size of a given depth (time $\times$ height $\times$ width)\\
        $\mathcal{D}_{cosine}$ & - & The cosine similarity distance function \\
        $\hat{\mathcal{D}}$ & - & The expanded diversity of cosine similarity distance function\\
        
    \bottomrule
    \end{tabular}
    }
    \label{tab:notation}
\end{table}

\section{Proof of Spatio-temporal Entropy Score}

\subsection{Derivation Process of Differential Entropy}
\label{ssec:proof entropy}

Suppose $x$ is sampled from Gaussian distribution $\mathcal{N} (\mu, \sigma^2)$, and we know about the probability density function of $x$:
\begin{equation}
    p(x) = \frac{1}{\sqrt{2\pi}\sigma}exp[-\frac{(x-\mu)^2}{2\sigma^2}]~.
    \label{eq:known1}
\end{equation}

We can then derive the differential entropy with $\int_{-\infty}^{+\infty}e^{-x^2}dx = \sqrt{\pi}$ as:
\begin{equation}
    \begin{aligned}
        \mathcal{H}(x)&= \int_{-\infty}^{+\infty}-log(p(x)p(x)dx \\
            &= -\int_{-\infty}^{+\infty}\frac{1}{\sqrt{2\pi}\sigma}exp[-\frac{(x-\mu)^2}{2\sigma^2}]log\frac{1}{\sqrt{2\pi}\sigma}exp[-\frac{(x-\mu)^2}{2\sigma^2}]dx\\
            &=\frac{log(\sqrt{2\pi}\sigma)}{\sqrt{\pi}}\int_{-\infty}^{+\infty}e^{-y^2}dy + \frac{1}{\sqrt{\pi}}\int_{-\infty}^{+\infty}e^{-y^2}y^2dy\\
            &=log(\sqrt{2\pi}\sigma) + \frac{1}{\sqrt{\pi}}\times[-\frac{1}{2}(0-\int_{-\infty}^{+\infty}e^{-y^2}dy)]\\
            &= \frac{1}{2}log(2\pi) + log(\sigma) + \frac{1}{2} ~~ \propto  log(\sigma^2)~.\\
    \end{aligned}
        \label{eq:gaussian proof}
\end{equation}

\subsection{Expectation and Variance of Feature Map }
\label{ssec:proof variance}

According to \textbf{Theorem 2} and \textbf{Theorem 3}, we can compute the expectation of $l^{th}$ layer feature map element ${\boldsymbol {x}}^l_{i}$, as:
\begin{equation}
\small
    \mathbb{E}({\boldsymbol {x}}^l_{i}) = \mathbb{E}(\sum_{t=1}^{K^{l}_t} \sum_{h=1}^{K^{l}_h} \sum_{w=1}^{K^{l}_w} \sum_{c=1}^{C^{l-1}} {\boldsymbol {x}}^{l-1}_{cthw} {\boldsymbol{W}}^l_{cthw}) = \sum_{t=1}^{K^{l}_t} \sum_{h=1}^{K^{l}_h} \sum_{w=1}^{K^{l}_w} \sum_{c=1}^{C^{l-1}} \Big[\mathbb{E}({\boldsymbol {x}}^{l-1}_{cthw}) \mathbb{E}({\boldsymbol{W}}^l_{cthw})\Big],
    \label{eq:expectation}
\end{equation}

Given two independent random variables $v_1$ and $v_2$, based on $\mathbb{D}(v) = \mathbb{E}(v^2) - \mathbb{E}(v)^2$ and \textbf{Theorem 2}, we can then calculate the variance of the product of these variables as:
% \begin{equation}
%     \sigma(v) = E(v^2) - E(v)^2
% \end{equation}
\begin{equation}
    \begin{aligned}
        \mathbb{D}^2(v_1v_2) &= \mathbb{E}(v_1^2 v_2^2) -\mathbb{E}(v_1v_2)^2 = \mathbb{E}(v_1^2)\mathbb{E}(v_2^2) - \mathbb{E}(v_1)^2\mathbb{E}(v_2)^2\\
        & = [\mathbb{D}(v_1) + \mathbb{E}(v_1)^2][\mathbb{D}(v_2)+\mathbb{E}(v_2)^2] - \mathbb{E}(v_1)^2\mathbb{E}(v_2)^2\\
        &= \mathbb{D}^2(v_1)\mathbb{D}^2(v_2)+\mathbb{D}^2(v_2){[\mathbb{E}(v_1)]^2}+\mathbb{D}^2(v_1){[\mathbb{E}(v_2)]^2}~,
    \end{aligned}
\end{equation}
We can then derive the variance of $\boldsymbol{x}_i^{l}$, based on \textbf{Theorem 2} and \textbf{Theorem 3}, as:
\begin{equation}
    \begin{split}
    \mathbb{D}^2({\boldsymbol {x}}^l_{i}) &= \mathbb{D}^2(\sum_{t=1}^{K^{l}_t} \sum_{h=1}^{K^{l}_h} \sum_{w=1}^{K^{l}_w} \sum_{c=1}^{C^{l-1}} {\boldsymbol {x}}^{l-1}_{cthw} {\boldsymbol{W}}^l_{cthw}) \\
    &= \sum_{t=1}^{K^{l}_t} \sum_{h=1}^{K^{l}_h} \sum_{w=1}^{K^{l}_w} \sum_{c=1}^{C^{l-1}} {\mathbb{D}^2(\boldsymbol {x}}^{l-1}_{cthw} {\boldsymbol{W}}^l_{cthw}) \\
    %  = \sum_{t=1}^{K^{l}_t} \sum_{h=1}^{K^{l}_h} \sum_{w=1}^{K^{l}_w} \sum_{c=1}^{C^{l-1}} {\mathbb{D}^2(\boldsymbol {x}}^{l-1}_{cthw} {\boldsymbol{W}}^l_{cthw})\\
     &=\sum_{t=1}^{K^{l}_t} \sum_{h=1}^{K^{l}_h} \sum_{w=1}^{K^{l}_w} \sum_{c=1}^{C^{l-1}}\Big\{\mathbb{D}^2({\boldsymbol{x}}^{l-1}_{cthw}) \mathbb{D}^2({\boldsymbol{W}}^l_{cthw})\\
    &\quad+\mathbb{D}^2({\boldsymbol{x}}^{l-1}_{cthw}){\Big[\mathbb{E}({\boldsymbol{W}}^l_{cthw})\Big]^2}+\mathbb{D}^2({\boldsymbol{W}}^l_{cthw}){\Big[\mathbb{E}({\boldsymbol{x}}^{l-1}_{cthw})\Big]^2}\Big\},\\
    % &={K^{l}_t}{K^{l}_h}{K^{l}_w}C^{l-1}\left[\sigma^2({\boldsymbol{x}}^{l-1}_{cthw}{\boldsymbol{W}}^l_{cthw})\right]\\
    % &=\sum_{t=1}^{K^{l}_t} \sum_{h=1}^{K^{l}_h} \sum_{w=1}^{K^{l}_w} \sum_{c=1}^{C^{l-1}}\sigma^2({\boldsymbol {x}}^{l-1}_{cthw}).\\
    % &={K^{l}_t}{K^{l}_h}{K^{l}_w}C^{l-1}\sigma^2({\boldsymbol {x}}^{l-1}_{cthw}).
    % &={K^{l}_t}{K^{l}_h}{K^{l}_w}C^{l-1}Var({\boldsymbol {x}}^{l-1}_{cthw}],\  with\ Var[{\bf {W}}^l_{cthw}]=1, 
    \end{split}
\end{equation}

\subsection{Proof of 3D CNNs Entropy}
\label{ssec:proof 3d cnn}
% In the first Conv layer, also suppose all parameters $\boldsymbol{W}$ in the spatio-temporal system are initialized from a standard Gaussian distribution, which means that $E({\boldsymbol{W}}^l_{cthw})=0$ and $\sigma^2({\boldsymbol{W}}^l_{cthw})=1$.

% \begin{equation}
% \label{eq:first}
% % \begin{aligned}
%     \mathbb{E}({\boldsymbol {x}}^1_{i}) =0~, \quad \mathbb{D}^2({\boldsymbol {x}}^1_{i}) = \sum_{t=1}^{K^{1}_t}\sum_{h=1}^{K^{1}_h} \sum_{w=1}^{K^{1}_w} \sum_{c=1}^{C^0} \Big[\mathbb{D}^2({\boldsymbol{x}}^0_{chw})\Big]~,
% \end{equation}

As the input $\boldsymbol{x}^0$ is initialized from a standard Gaussian distribution $\mathcal{N}(0,1)$, and all parameters initialized from Gaussian Distribution  $\mathcal{N}(0,\sigma^2_w)$, we can formulate Eq. (\ref{eq:expectation}) and Eq. (\ref{eq:variance}) as:
\begin{equation}
\label{eq:first2}
% \begin{aligned}
    \mathbb{E}({\boldsymbol {x}}^1_{i}) =0, \quad \mathbb{D}^2({\boldsymbol {x}}^1_{i}) = \sum_{t=1}^{K^{1}_t}\sum_{h=1}^{K^{1}_h} \sum_{w=1}^{K^{1}_w} \sum_{c=1}^{C^0} \Big[\mathbb{D}^2({\boldsymbol{W}}^1_{chw})\Big],
\end{equation}
Subsequently, the expectation $\mathbb{E}({\boldsymbol {x}}^L_{i})$ and variance $\mathbb{D}^2(x^L)_{i}$ of the last layer can be derived as:
\begin{equation}
\label{eq:next2}
    \mathbb{E}({\boldsymbol {x}}^L_{i}) =0, \quad \mathbb{D}^2({\boldsymbol {x}}^L_{i}) = \sum_{t=1}^{K^{L}_t}\sum_{h=1}^{K^{L}_h} \sum_{w=1}^{K^{L}_w} \sum_{c=1}^{C^{L-1}} \Big[\mathbb{D}^2({\boldsymbol{W}}^L_{chw})\Big],
% \end{aligned}
\end{equation}
Therefore, the variance can be computed by propagating the variances from previous layers as:
\begin{equation}
    \mathbb{D}^2({\boldsymbol {x}}^L_{i}) = \prod^{L}_{l=1}{{K^{l}_t}{K^{l}_h}{K^{l}_w}C^{l-1}}\mathbb{D}^2({\boldsymbol{W}^l_{chw}}),
\end{equation}
According to Eq. (\ref{eq:gaussian}), the upper bound entropy is proportional to the variance of last feature map. Then we can derive Eq. (\ref{eq:gaussian}) as:
\begin{equation}
    \mathcal{H}(F)  \propto  \sum^{L}_{l=1}log({{K^{l}_t}{K^{l}_h}{K^{l}_w}C^{l-1}}\mathbb{D}^2({\boldsymbol{W}^l_{chw}}))~,\\
\end{equation}

\newpage
\section{Discussion of Simple Network Space}
\label{sec:simple network}

The bias of a convolutional layer is zero, and the activation function in the network is omitted in the search for simplification,  following the work of ZenNAS \citep{lin2021zen} and MAE-DET \citep{sun2022mae}, which has been shown to have no influence on the expressiveness of the network.    
The training of CNN models has been well studied, and some components can be integrated to boost performance. We deliberately avoid using these components to keep our design simple and universal. Nevertheless, these auxiliary components can easily be plugged into the architecture without any special modifications.
Moreover, we provide a discussion of auxiliary components with the entropy calculation, which is listed below.

\noindent\textbf{Batch Normalization (BN)}.
BN is a widely used method to re-center and re-scale the features to make the network converge faster and more stable. BN normalizes entropies adaptively to the network width (which can be related to output variance). When BN is used, networks of different widths will have the same entropy value. Hence, BN has to be removed when calculating entropy.

\noindent\textbf{Activation Function}. 
Activation functions increase the non-linearity of training, which has different effects on entropy. For example, ReLUs, half the variance of the output, decrease the entropy with a constant factor in each layer, having a less positive effect on entropy. Meanwhile, if we formulate each kind of activation for our system, it introduces redundancy and becomes complicated, so we give them a uniform form to omit them in search of concise expressiveness calculation. 
% All activation functions, especially ReLU will change the value of feature map in each depth, which will affect the calculation of expectation and variance. Hence, activation functions has to be removed when calculating the entropy.

\noindent\textbf{Residual Link}.
If the input and all parameters are initialized from standard Gaussian distribution, the variances with or without residual links are less than 2\% different in entropy score, which means it affects the entropy value slightly. Meanwhile, the residual link has a significant impact on convergence in training.

\noindent\textbf{Squeeze-and-Excitation Module (SE)}.
SE modules are used to adaptively recalibrate channel-wise feature responses by explicitly modeling interdependency between channels. When the input is initialized from a Gaussian distribution, the output after global pooling in SE block is equal to 0 and the final output becomes 0.5, which will lose the ability to model interdependency between channels.

\section{Comparison on Training-Free Scores}
\label{sec:entropy comparison}

\subsection{Comparison on the ImageNet-1K dataset.}
\begin{table}[h]
\captionsetup{font={small}}
    \centering
    % \resizebox{\textwidth}{!}{
    \begin{tabular}{ccccc}
    \toprule
       Training-free method   & FLOPs & \begin{tabular}[c]{@{}c@{}}Search \\ Devices\end{tabular} & \begin{tabular}[c]{@{}c@{}}Design Cost\\ (hours)\end{tabular} & Top-1 \\
    \midrule
       ResNet-50  & 4.1G & - & - & 78.0 \\
       \midrule
       Zen-score \citep{lin2021zen}  & 4.4G & GPU$\ddag$ & 24 & 78.9 \\
       MAE-DET score \citep{sun2022mae}  & 4.4G &  GPU$\ddag$ & 14  & 79.1\\
       \midrule
       HomoEntr-Score (w/o K$^t$) & 4.3G & CPU$\dag$ & 3  & 79.0 \\
    \bottomrule
    \end{tabular}
    \caption{Comparison of different training-free methods on ImageNet-1K dataset. $\ddag$: Nvidia Tesla V100 16G GPU, $\dag$: AMD Ryzen 5 5600X 6-Core CPU.}
    \label{tab:training-free comparison}
    % \vspace{-0.5cm}
\end{table}

% \textcolor{blue}{lack of search cost comparison, which is our advantage on efficiency. such as zen 12 GPU hours, mae-det 14 GPU hours, STE-standard 0.5 CPU hour.}
We conduct comparison experiments using 2D CNNs on ImageNet with the same evolutionary strategies (ResNet design space), as shown in Table \ref{tab:training-free comparison}. 
Compared with the result of ResNet-50, the model searched by HomoEntr-Score improves 1.0\% of accuracy, which indicates that the entropy-based analytic formulation can also measure the information capacity of 2D CNNs.
Compared with Zen-score and MAE-DET, the performance of our proposed formulaic metric can also achieve comparable performance. It means that HomoEntr-Score can work well for modeling the information capacity of 2D CNNs, since there is no (obvious) discrepancy in the information of the two directions in 2D images statistically.

\subsection{Comparison on the Sth-Sth V1 dataset.}
Since there is no existing code available for training-free NAS methods for 3D CNNs, we then refine their implementations for the video recognition task. The results are shown in Table \ref{tab:training-free comparison 3D}.
    \begin{table}[h]
    \centering
    % \resizebox{0.75\textwidth}{!}{
    \begin{tabular}{cccccc}
    \toprule
       Training-free method   & GFLOPs & \begin{tabular}[c]{@{}c@{}}Search \\ Devices\end{tabular} & \begin{tabular}[c]{@{}c@{}}Design Cost\\ (hours)\end{tabular} & Top-1 & Top-5\\
    \midrule
          X3D-S \citep{feichtenhofer2020x3d} & 2G & - & - & 44.6 & 74.4 \\
       \midrule
       Zen-score \citep{lin2021zen} & 1.9G & GPU & 26 & 45.5 & 74.6\\
       MAE-DET score \citep{sun2022mae}  & 1.9G &  GPU& 15  & 45.8 & 74.7\\
      \midrule
       E3D-S & 1.9G & CPU & 3  & 47.1 & 75.6 \\
    \bottomrule
    \end{tabular}
    \caption{Comparison of different training-free methods on the Sth-Sth V1 dataset. $\ddag$: Nvidia Tesla V100 16G GPU, $\dag$: AMD Ryzen 5 5600X 6-Core CPU.}
    \label{tab:training-free comparison 3D}
    % \vspace{-0.5cm}
\end{table}

According to the results in Table \ref{tab:training-free comparison 3D}, the performances of other training-free NAS methods are better than X3D-S, but the performance of our searched model is higher.
It indicates directly applying training-free NAS methods can be effective in the video recognition task, but it still needs spatio-temporal refinement on video understanding tasks, which our work mainly focuses on.

\section{Detailed Searching Algorithm and Settings}
\label{sec:algorithm}
To obtain highly expressive 3D CNNs of maximized entropy, we use a customized Evolutionary Algorithm. The step-by-step description of EA is given in Algorithm \ref{al:ea}, as the architecture generator.
We only apply the STEntr-Score to guide the evolution process, not accuracy, which therefore does not need training on the dataset.
We choose EA due to its simplicity, and it is possible to choose other methods, such as reinforcement learning or even greedy selection.
According to our kernel selection observations, we define the 3D kernel size search space within each layer, $1\times{(k^{space})^2}$, $k^{times} \times (k^{space})^2$, to be chosen as one of the following: \{1$\times$3$\times$3, 1$\times$5$\times$5, 3$\times$3$\times$3\}. These choices enable a layer to focus on and aggregate different dimensional representations efficiently, expanding the network’s receptive field in the most pertinent directions, while reducing FLOPs along other dimensions \citep{kondratyuk2021movinets}.
% Due to different kernel size may benefit from having different numbers of input filters, we search over a range of bottleneck widths as multipliers in $\{1.5, 2.0, 2.5, 3.0, 3.5, 4.0\}$ and output channel widths as multipliers in $\{2.0, 1.5, 1.25, 0.8, 0.6, 0.5\}$.
% As we apply MobileNet-like network basis, each layer surrounds the 3D convolution with two $1\times1^2$ convolutions to expand.

\subsection{Initial Architecture}
\begin{table}[ht]
% \begin{wraptable}{r}{6cm}
    \centering
    \caption{E3D-S initial searching. ``stage{$_x$}" is a super structure which contains ``layers"-layer 3D inverted bottleneck block. Channels means the output channels of the corresponding convolution.}
        \renewcommand\arraystretch{1.05}
        \setlength\tabcolsep{6pt}
        % \resizebox{\textwidth}{!}{
        \scalebox{1}{
        \begin{tabular}{ccccc}
    	    \toprule
    		Stage & Kernels & Channels & Layers & $T \times H \times W$\\
    		\midrule
    		\multirow{1}{*}{data} & \multirow{1}{*}{stride 6, 1$^\text{2}$} & 3 & 1 &  $13 \times 160 \times 160$   \\
    		%			&  &  \\
    		\midrule
    		\multirow{1}{*}{conv$_1$} & \multirow{1}{*}{$1\times3^\text{2}$, {24}}  & 24 & 1 & $13 \times 80 \times 80$  \\
    		%			& &  \\
    		\midrule
    		{stage$_2$}  & [1$\times$1$^\text{2}$, 3$\times$3$^\text{2}$, 1$\times$1$^\text{2}$] & [48,\ 48,\ 24] & 1 & {$13 \times 40 \times 40$}  \\
    % 		\hline
    		{stage$_3$}  & [1$\times$1$^\text{2}$, 3$\times$3$^\text{2}$, 1$\times$1$^\text{2}$] & [96,\ 96,\ 48] & 1 & {$13 \times 20 \times 20$} \\
    % 		\hline
    		{stage$_4$}  & [1$\times$1$^\text{2}$, 3$\times$3$^\text{2}$, 1$\times$1$^\text{2}$] & [192,\ 192,\ 96] & 1 & {$13 \times 10 \times 10$} \\
    % 		\hline
    		{stage$_5$}  & [1$\times$1$^\text{2}$, 3$\times$3$^\text{2}$, 1$\times$1$^\text{2}$] & [192,\ 192,\ 96] & 1 & {$13 \times 10 \times 10$} \\
    % 		\hline
    		{stage$_6$}  & [1$\times$1$^\text{2}$, 3$\times$3$^\text{2}$, 1$\times$1$^\text{2}$] & [384,\ 384,\ 192] & 1 & {$13 \times 5 \times 5$}  \\
    		\midrule
    		\multirow{1}{*}{conv$_7$} & \multirow{1}{*}{$1\times1^\text{2}$}  & 512 & 1 & $13 \times 5 \times 5$   \\
    		%			&  &  \\
    		\multirow{1}{*}{pool$_8$} & $13 \times 5 \times 5$ & 512 &    1 & $1 \times 1 \times 1$    \\
    		%			&  &  \\
    		\multirow{1}{*}{conv$_{9/10}$} & \multirow{1}{*}{$1\times1^\text{2}$, $1\times1^\text{2}$}   & [2048,\  \#classes] & 1 & $1 \times 1 \times 1$    \\
    		%			&  &  \\
    % 		\multicolumn{1}{c}{conv$_{10}$}  & \multirow{1}{*}{$1\times1^\text{2}$}  & { \#classes}& 1 & $1 \times 1 \times 1$  \\
    		\bottomrule
        \end{tabular}}
        \label{tab:init}
% \end{wraptable}
\end{table}

Firstly, we set up the initial architecture in a MobileNet-styled network, as shown in Table \ref{tab:init}, which consists of five stages with only one layer that can be easily evolved during the algorithm. The initial architecture is inspired by the structure of X3D-S \citep{feichtenhofer2020x3d} because inheriting good prior design can reduce the uncertainty of search space.
Then, based on the initial architecture, applied EA helps us mutate channel dimension, kernel selection, bottleneck expansion ratio, and layer arrangement by randomly selecting the stage. Note that the channel dimension in conv$_1$ and conv$_7$ also participate in the mutation process.
% Note that the applied Evolutionary algorithm only help us mutate kernel selection, channel dimension (from conv$_1$ to conv$_7$), and layer arrangement (from stage$_2$ to stage$_6$), and won't affect the basic architecture itself.

\begin{algorithm}[ht]
 \caption{Maximum Entropy Evolutionary Algorithm}
 \label{ag:evolution}
 	\begin{algorithmic}[1]
		
		\REQUIRE Search space $\mathcal{S}$. Inference budget $B$, maximal depth $L$, total number of iterations $M$, evolutionary population size $N$, initial structure $F_0$.
		
		\ENSURE Designed E3D backbone $F^*$.
		
		\STATE Initialize population $\mathcal{P}=\{F_0\}$.
		\FOR{$m=1,2,\cdots,M$}
		\STATE Randomly select $F_m \in \mathcal{P}$ and select two stages $stage_k \in F_m$.
		
		\FOR{$j=1,2$}
		\STATE {\bf Switch}\ {Randomly select one target of \{{\bf Kernel} size, {\bf Output} channels, {\bf Bottleneck} channels, {\bf Layers}\} from $stage_{kj}$}\ {\bf do}
		\STATE {\bf Case kernel:}\ {Mutate kernel from 3D kernel search space.}
		\STATE {\bf Case Output:}\ {Mutate output channels with multiplier space.}
		\STATE {\bf Case Bottleneck:}\ {Mutate bottleneck channels with expansion ratio space.}
		\STATE {\bf Case Layers:}\ {Mutate block layers with addend from $\{-2, -1, 1, 2\}$.}
		\ENDFOR
		\STATE Get mutated network $\hat{F}_m$ with two mutated stages $stage_{kj}$.		
		\IF{$\hat{F}_m$ is within inference budget $B$ and has no more than $L$ layers}
		\STATE Get STEntr-Score of $\hat{F}_m$ and append $\hat{F}_m$ to $\mathcal{P}$.
		\ENDIF
		\STATE Remove networks of the smallest STEntr-Score if the size of $\mathcal{P}$ exceeds $B$.
		\ENDFOR
		\STATE Return $F^*$, the network of the highest STEntr-Score in $\mathcal{P}$.
		
	\end{algorithmic}
	\label{al:ea}
\end{algorithm}

\subsection{Evolutionary Algorithm}

In Algorithm \ref{al:ea}, we randomly initialize a population of candidates from the initial structure, under a computational budget. The population size and total iterations of EA are set to 512 and $500000$, respectively.
% with layer arrangement (1-\textcolor{red}{85}).
At each iteration step $m$, we randomly select two stages from the candidates and mutate them. Next, we will randomly select a mutation strategy from 4 strategies for each stage.
Specific mutation strategies for our E3D family are described as follows. We randomly select 3D kernels from \{1$\times$3$\times$3, 1$\times$5$\times$5, 3$\times$3$\times$3\} to replace the current one; interchange the expansion ratio of bottleneck from $\{1.5, 2.0, 2.5, 3.0, 3.5, 4.0\} $($bottleneck = ratio\times intput$); scale the output channels with the ratios $\{2.0, 1.5, 1.25, 0.8, 0.6, 0.5\}$; or increases or decreases depth with 1 or 2. 
% These mutation strategies will assure the diversity of variants. 
Note that the channel dimension of every layer is fixed within from 8 to 640 with multiples of 8, which will help shrink homologous search space and accelerate the search speed.
The mutated structure $\hat{F}_m$ is appended to the population if its inference cost does not exceed the budget.
Finally, we maintain the population size by removing networks with the smallest STEntr-Score.
After $M$ iterations, the target network with the largest STEntr-Score is obtained, namely E3D.

\section{E3D Family Architecture Details}
\label{sec:architecture}

Table \ref{tab:family} shows three instantiations of E3D with varying complexity, including E3D-S (1.9G FLOPs), E3D-M (4.7G FLOPs), and E3D-L (18.3G FLOPs).
All models are searched separately with different FLOPs budget (1.9G, 4.7G, and 18.4G) for a fair comparison with X3D-S/M/L as the baseline.
Meanwhile, SE block and ReLU activation function will be added into these architectures for training.
For both training and inference, the input size remains the same: 160 for E3D-S, 224 for E3D-M, and 312 for E3D-L. 
All channel dimensions and layer arrangements are searched by evolutionary algorithm under different given budgets.

\begin{table}[ht]
    \centering
    \resizebox{\textwidth}{!}{
    \begin{tabular}{c cc cc cc}
    \toprule
    \multirow{2}{*}{Stage} &
    \multicolumn{2}{c}{E3D-S}&
    \multicolumn{2}{c}{E3D-M}&
    \multicolumn{2}{c}{E3D-L} \\
    % \cmidrule(cc){2-3}
    % \cmidrule(cc){4-5}
    % \cmidrule(cc){6-7}
    \multicolumn{1}{c}{} &
    \multicolumn{1}{c}{filters} &
    \multicolumn{1}{c}{output size}   &
    \multicolumn{1}{c}{filters} &
    \multicolumn{1}{c}{output size}   &
    \multicolumn{1}{c}{filters} &
    \multicolumn{1}{c}{output size}  \\
    \midrule
      data   & stride 6, 1$^\text{2}$ & $13 \times 160 \times 160$ & stride 5, 1$^\text{2}$& $16 \times 224 \times 224$ & stride 5, 1$^\text{2}$ & $16 \times 312 \times 312$ \\
    \midrule
       conv$_1$  & $1\times3^\text{2}$, 24 & $13 \times 80 \times 80$ & $1\times3^\text{2}$, 24 & $16 \times 112 \times 112$ & $1\times3^\text{2}$, 24 & $16 \times 156 \times 156$\\ 
       \midrule
       stage$_2$ & ${\begin{bmatrix} 1\times1^\text{2}, 32 \\ 1\times5^\text{2}, 32 \\ 1\times1^\text{2}, 24 \end{bmatrix}}\times$3 & $13 \times 40 \times 40$ &${\begin{bmatrix} 1\times1^\text{2} , 32 \\ 1\times5^\text{2} , 32 \\ 1\times1^\text{2}, 24 \end{bmatrix}}\times$3 & $16 \times 56 \times 56$ & ${\begin{bmatrix} 1\times1^\text{2}, 32 \\ 1\times5^\text{2}, 32 \\ 1\times1^\text{2}, 24 \end{bmatrix}}\times$3 & $16 \times 78 \times 78$ \\
       
       stage$_3$ & ${\begin{bmatrix} 1\times1^\text{2}, 96 \\ 3\times3^\text{2}, 96 \\ 1\times1^\text{2}, 48 \end{bmatrix}}\times$6 & $13 \times 20 \times 20$
       & ${\begin{bmatrix} 1\times1^\text{2}, 96  \\ 3\times3^\text{2}, 96 \\ 1\times1^\text{2}, 64 \end{bmatrix}}\times$6  & $16 \times 28 \times 28$ & ${\begin{bmatrix} 1\times1^\text{2}, 120 \\ 3\times3^\text{2}, 120 \\ 1\times1^\text{2}, 48 \end{bmatrix}}\times$13 & $16 \times 39 \times 39$\\
       
       stage$_4$ & ${\begin{bmatrix} 1\times1^\text{2}, 176 \\ 3\times3^\text{2}, 176 \\ 1\times1^\text{2}, 120 \end{bmatrix}}\times$6 & $13 \times 10 \times 10$ &  ${\begin{bmatrix} 1\times1^\text{2}, 176  \\ 3\times3^\text{2}, 176 \\ 1\times1^\text{2}, 120 \end{bmatrix}}\times$6  & $16 \times 14 \times 14$ & ${\begin{bmatrix} 1\times1^\text{2}, 176 \\ 3\times3^\text{2}, 176 \\ 1\times1^\text{2}, 120 \end{bmatrix}}\times$13& $16 \times 20 \times 20$\\
       
       stage$_5$ & ${\begin{bmatrix} 1\times1^\text{2}, 176 \\ 3\times3^\text{2}, 176 \\ 1\times1^\text{2}, 120 \end{bmatrix}}\times$6 & $13 \times 10 \times 10$ &  ${\begin{bmatrix} 1\times1^\text{2}, 176  \\ 3\times3^\text{2}, 176 \\ 1\times1^\text{2}, 120 \end{bmatrix}}\times$6  & $16 \times 14 \times 14$ & ${\begin{bmatrix} 1\times1^\text{2}, 176 \\ 3\times3^\text{2}, 176 \\ 1\times1^\text{2}, 120 \end{bmatrix}}\times$13& $16 \times 20 \times 20$\\
       
       stage$_6$ & ${\begin{bmatrix} 1\times1^\text{2}, 384 \\ 3\times3^\text{2}, 384 \\ 1\times1^\text{2}, 256 \end{bmatrix}}\times$6 & $13 \times 5 \times 5$ & ${\begin{bmatrix} 1\times1^\text{2}, 464  \\ 3\times3^\text{2}, 464 \\ 1\times1^\text{2}, 184 \end{bmatrix}}\times$6  & $16 \times 7 \times 7$ & 
       ${\begin{bmatrix} 1\times1^\text{2}, 480 \\ 3\times3^\text{2}, 480 \\ 1\times1^\text{2}, 192 \end{bmatrix}}\times$13 &  $16 \times 10 \times 10$\\
       \midrule
       conv$_7$ & $1\times1^\text{2}$,  & $13 \times 5 \times 5$ & $1\times1^\text{2}$, 464 & $16 \times 7 \times 7$ & $1\times1^\text{2}$, 480 & $16 \times 10 \times 10$\\
       pool$_8$ & $13 \times 5 \times 5$ & $1 \times 1 \times 1$ &$16 \times 7 \times 7$ & $1 \times 1 \times 1$ & $16 \times 10 \times 10$ & $1 \times 1 \times 1$\\
       conv$_{9/10}$ & [2048, \#classes] & $1 \times 1 \times 1$&[2048, \#classes] &$1 \times 1 \times 1$ & [2048, \#classes]& $1 \times 1 \times 1$\\
    \bottomrule
    \end{tabular}}
    \caption{Three instantiations of E3D with varying complexity. E3D-S with 1.9G FLOPs, E3D-M with 4.7G FLOPs, and E3D-L with 18.4G FLOPs. The size of output is $T \times H \times W$.}
    \label{tab:family}
\end{table}

\section{Experiment Setting Details}
\label{sec:setting}

\subsection{Datasets}
Our experiments are conducted on three large-scale datasets: Something-Something (Sth-Sth) V1\&V2 \citep{goyal2017something}, and Kinetics400 \citep{kay2017kinetics}. More dataset details can be seen in the supplementary materials.
1) The Sth-Sth datasets are more focused on fine-grained and motion-dominated actions, which contain pre-defined basic actions involving different interacting objects. 
Sth-Sth V1 comprises 86k video clips in the training set and 12k video clips in the validation set. Sth-Sth V2 is an updated version of Sth-Sth V1, which contains 169k video clips in the training set and 25k video clips in the validation set. They both have 174 action categories.
2) The Kinetics dataset contains activities in daily life and some categories are highly correlated with interacting objects or scene context.
Kinetics400 contains over 200k training videos and 20k validation videos divided into 400 categories, covering a wide range of human activities.

\subsection{Implementation Details}
Detailed implementation settings of \textbf{training \& inference stage} on Sth-Sth V1\&V2 and Kinetics400 datasets are listed in Table \ref{tab:implementation}.
All experiments are performed on 8$\times$Nvidia Tesla A100 GPUs.

% \newpage
\begin{table}[ht]
    \centering
    \begin{tabular}{l c c}
    \toprule
    Hyperparameter & Sth-Sth V1\&V2 & Kinetics400 \\
    \midrule
       Epoch  & 128 & 256 \\
    Batch Size per GPU & 32 & 16\\
    Optimizer     & SGD & SGD\\
    Learning Rate & 0.8 & 0.4 \\
    Learning Rate Policy & cosine & cosine\\
    Momentum & 0.9 & 0.9 \\
    Weight Decay & 5e$^{-5}$ & 5e$^{-5}$\\
    Warm-up Epoch & 10 & 15 \\
    Synchronized Batch Normalization & True & True \\
    Training from scratch & True & True \\
    \bottomrule
    \end{tabular}
    \caption{List of hyperparameters used on Sth-Sth V1\&V2 and Kinetics400 datasets.}
    \label{tab:implementation}
\end{table}

% \newpage
\section{Additional Results}
\label{sec:more results}

% \subsection{Multi-view Results}
\subsection{Accuracy vs. Complexity}
\begin{wrapfigure}{r}{6.5cm}
    \centering
    % \fbox{\rule{0pt}{1.5in} \rule{0.9\linewidth}{0pt}}
    \includegraphics[width=0.95\linewidth]{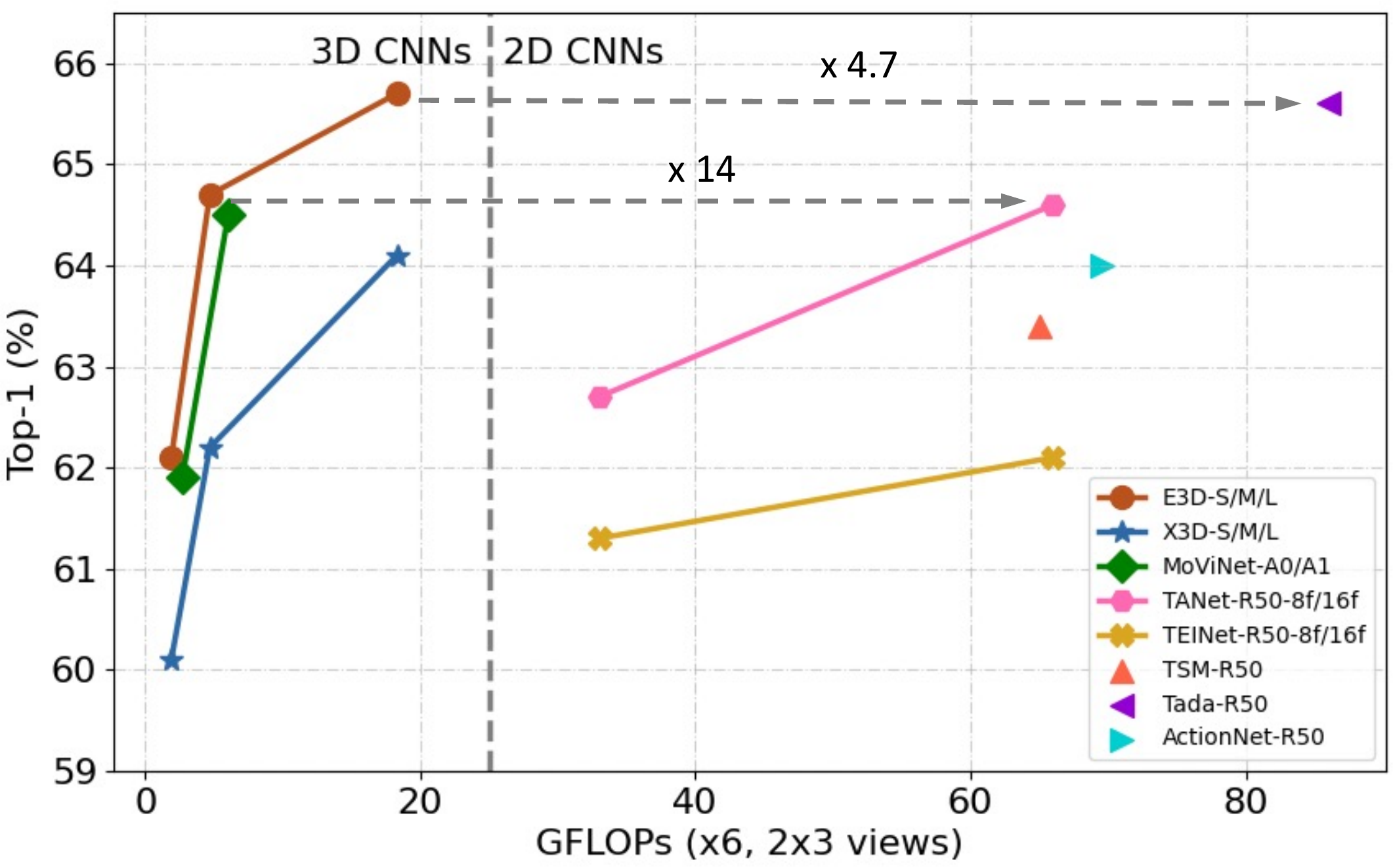}
    \caption{Accuracy/complexity trade-off on the Sth-Sth V2 dataset.}
    \label{fig:tradoff}
\end{wrapfigure}
% For the Sth-Sth V2 dataset, the inference procedure follows a fixed number of clips for testing (2 clips $\times$ 3 crops).
% Figure \ref{fig:tradoff} shows the trade-off for the full inference of a video, when varying the number of temporal clips used.
% Each model demonstrates a large improvement when changing $K=1$ clip to $K=2$ clips testing, which doubles the FLOPs.
% Compared to TANet, although performances are similar, 2D CNNs based methods will consume more computational resources.
% Compared to X3D-M, we observe that both E3D and MoViNet can achieve large improvement, which indicates that searched methods have higher efficiency in utilizing computing resources. 
% Also, our method achieves comparable performance compared with MoViNet, which indicates that the proposed training-free STEntr-Score can effectively evaluate the expressiveness of a 3D architecture.
Figure \ref{fig:tradoff} shows the trade-off between accuracy and complexity (FLOPs). 
Compared to 2D CNN-based methods, E3D requires much lower computational resources. 
Although the performance of our method is similar to Tada-R50, the FLOPs of Tada-R50 are 4.7 times more than E3D-L.
Compared to 3D CNN-based methods, we observe that both E3D and MoViNet can achieve large improvement, which indicates that searched methods have higher efficiency in utilizing computing resources. Also, our method achieves comparable performance compared with MoViNet, which indicates that the proposed training-free STEntr-Score can effectively evaluate the expressiveness of a 3D architecture.

\subsection{HomoEntr-Score vs. STEntr-Score}
Table \ref{tab:hescore vs stescore} reports E3D results searched by HomoEntr-Score and STEntr-Score, under the same search settings.
The results show substantial improvement when using STEntr-Score instead of HomoEntr-Score, which indicates the effectiveness of STEntr-Score to handle the discrepancy of visual information in spatial and temporal dimensions.
Even though without refinement factor, the performance of HomoEntr-Score searched E3D still outperforms X3D, which means the entropy-based search strategy can also 
measure the expressiveness of 3D CNN architectures.

\begin{table}[h]
\captionsetup{font={small}}
    \centering
    % \resizebox{\textwidth}{!}{
    \begin{tabular}{lcccc}
    \toprule
       Model   & Resolution & GFLOPs & Top-1 & Top-5\\
    \midrule
      X3D-S* \citep{feichtenhofer2020x3d} & 13$\times$160$^2$ & 2 & 44.6 & 74.4\\
      MoViNet-A0* \citep{kondratyuk2021movinets} & 50$\times$172$^2$ & 2.7 &  46.9 & 75.0 \\
    \midrule
      E3D (HomoEntr-Score) & 13$\times$160$^2$ & 1.9 &  45.8 & 74.8\\
      E3D (STEntr-Score) & 13$\times$160$^2$ & 1.9 &  47.1 & 75.6\\
    \bottomrule
    \end{tabular}
    \caption{Comparison of different entropy scores on the Sth-Sth V1 dataset. * denotes our reproduced models.}
    \label{tab:hescore vs stescore}
    % \vspace{-0.5cm}
\end{table}

% \newpage
% \subsection{Search Strategy Ablation Study}
\subsection{3D Kernel Search Space}
% \noindent\textbf{3D Kernel Search Space}.
To analyze the impact of kernel search space, we expand the 3D kernel search space and conduct experiments, as shown in Table \ref{tab:search space}.
The results indicate that larger search spaces actually benefit the performance. However, compared to the results between E3D (HomoEntr-Score) with E3D (STEntr-Score)) in Table \ref{tab:hescore vs stescore}, the STEntr-Score based searching can boost the performance (+1.3\%) more than a large search space did (+0.2\%). 
It also verified the effectiveness of our proposed STEntr-Score in evaluating the expressiveness of 3D CNNs.

\begin{table}[h]
\captionsetup{font={small}}
    \centering
    % \resizebox{0.8\textwidth}{!}{
    \begin{tabular}{cccc}
    \toprule
       Kernel Search Space   & FLOPs & Top-1 & STEntr-Score\\
    \midrule
      1$\times$3$\times$3, 3$\times$3$\times$3  & 1.9G &  46.3 & 198.55\\
      1$\times$3$\times$3, 1$\times$5$\times$5, 3$\times$3$\times$3  & 1.9G &  47.1 & 202.86\\
      1$\times$3$\times$3, 1$\times$5$\times$5, 3$\times$3$\times$3, 3$\times$1$\times$1  & 1.9G &  47.1 & 202.74\\
      1$\times$3$\times$3, 1$\times$5$\times$5, 3$\times$3$\times$3, 5$\times$3$\times$3  & 1.9G &  47.2 & 203.13\\
    %   1$\times$3$\times$3, 1$\times$5$\times$5, 3$\times$3$\times$3, 5$\times$3$\times$3, 3$\times$1$\times$1  & 1.9G &  47.1 & 203.01\\
    %   [3$\times$3$\times$3, 1$\times$5$\times$5]  & 1.9G &  79.0 &\\
    \bottomrule
    \end{tabular}
    \caption{Comparison of different 3D kernal search space on the Sth-Sth V1 dataset.}
    \label{tab:search space}
    % \vspace{-0.5cm}
\end{table}
\newpage
\subsection{Inference time comparison}
We report the inference time comparison with some state-of-the-art methods in Table \ref{tab:time}.
All models are trained and tested on the Sth-Sth V1 dataset, and the batch size is set to 16.
Compared to X3D, our E3D performs better not only on accuracy but also costs lower inference time. It indicates that the searched architecture by our proposed STEntr-Score is more effective and efficient for video understanding.
Compared to MoViNet, even though Top-1 accuracies are similar, both latency and throughput of E3D are performing better. Due to MoViNet applies a causal convolutional network and contains more parameters.
Compared to 2D CNN-based methods, E3D performs better on both accuracy and running time and requires much lower computational resources.
Overall, we believe that our proposed E3D family is more efficient and practical for real-world applications.

\begin{table}[ht]
    \caption{Inference comparison using a Tesla V100 on the Sth-Sth V1 dataset.}
    \centering
    \setlength\tabcolsep{3pt}
    \resizebox{\textwidth}{!}{
    \begin{tabular}{lccccccc}
    	     \toprule
    		 Method & Resolution & Frame  & GFLOPs & \#Param & Top1 & Latency (ms/video) & Throughput(video/s)\\
    		\midrule
    		TSM \citep{lin2019tsm} & 256 & 16 & 65 & 23.9M & 47.2 & 23.0 & 43.5\\
    		TANet \citep{liu2021tam} & 256 & 16  & 66 & 26M & 47.6 & 14.7 & 68.0\\
    		\midrule
    % 		X3D-S \citep{feichtenhofer2020x3d} & 160 & 13  & 2 & 3.7M & 44.6 &  & \\
    		X3D-M \citep{feichtenhofer2020x3d} & 224 & 16  & 4.7 & 3.7M & 47.3 & 13.5 & 74.1\\
    		MoViNet-A1 \citep{kondratyuk2021movinets} & 172 & 50  & 6 & 4.6M & 49.3 & 21.9 & 45.7\\
    		
    		\midrule
		  %  E3D-S & 160 & 13  & 1.9 & 3.4M & 47.1 &  & \\
		    E3D-M & 224 & 16  & 4.7 & 3.4M & 49.4 & 11.4 & 87.7\\
    		\bottomrule[1pt]
    		
        \end{tabular}}
    
    \label{tab:time}
\end{table}

\section{Future Direction}
% \section{\textcolor{blue}{Limitation Discussion}}
\label{sec:future}

% Data distribution is powerful for ranking the architectures. 
% However, the process of STEntr-Score search is contained without data training.
\noindent\textbf{Data-driven design}.
The design of STEntr-Score search correlates with parameter initialization and kernel selection, with standard Gaussian initialization input.
If we replace the Gaussian input directly with target data, the output after a convolution will be random due to the Gaussian initialized weights, as the process of STEntr-Score based searching is contained without data training. 
The aim of our work is therefore to provide a training-free approach to 3D CNN architecture design according to the maximum entropy principle under the given budgets. We believe that the training-free method, combined with target data without training, could be a future direction for research.

\noindent\textbf{Transformer model}.
We believe that the principle of maximum entropy is theoretically applicable to transformers. However, there exist some challenges to overcome. For example, Transformer has more complex components than CNN, such as `Q' and `K' kernel operation and multi-head attention, which is difficult to calculate the maximum entropy. In addition, the discrepancy of visual information in spatial and temporal dimensions by Transformer still remains a challenge. Although these challenges are difficult to overcome, this would be a fascinating task for us in the future.

\end{document}